\documentclass[10pt,twocolumn,letterpaper]{article}

\usepackage[pagenumbers]{cvpr} %

\usepackage[export]{adjustbox}
\usepackage{booktabs} %
\usepackage{graphicx}

\usepackage{tikz}

\usepackage{colortbl}
\usepackage{graphicx}
\usepackage{subcaption}
\usepackage{xspace}
\usepackage{multirow}
\usepackage{makecell}
\usepackage{algorithm}
\newcommand{\best}{\cellcolor{tablered}}
\newcommand{\sbest}{\cellcolor{orange}}
\newcommand{\tbest}{\cellcolor{yellow}}

\makeatletter
\DeclareRobustCommand\onedot{\futurelet\@let@token\@onedot}
\def\@onedot{\ifx\@let@token.\else.\null\fi\xspace}

\makeatother

\definecolor{yellow}{rgb}{1, 1, 0.7}
\definecolor{orange}{rgb}{1, 0.85, 0.7}
\definecolor{tablered}{rgb}{1, 0.7, 0.7}
\definecolor{red}{rgb}{1, 0, 0}

\definecolor{wincolor}{rgb}{0.85, 0.0, 0.0}

\definecolor{darkyellow}{rgb}{0.8, 0.8, 0.5}
\definecolor{darkred}{rgb}{0.7, 0.3, 0.3}
\definecolor{darkgreen}{rgb}{0.3, 0.7, 0.3}
\definecolor{green}{rgb}{0, 1.0, 0}
\definecolor{pink}{rgb}{1, 0.4, 0.7}

\definecolor{cvprblue}{rgb}{0.21,0.49,0.74}
\usepackage[pagebackref,breaklinks,colorlinks,allcolors=cvprblue]{hyperref}

\def\confName{CVPR}
\def\confYear{2026}

\title{Underwater360: Reconstructing Underwater Scenes from Panoramic Images with Omnidirectional Gaussian Splatting}

\author{Jiangbei Hu$^{1*}$\quad Weichao Song$^{1}$\thanks{Equal contribution}\quad Shibo Yu$^{1}$\quad Mohan Wang$^{1}$\quad Zihan Yi$^{1}$\quad Rui Wu$^{1}$\\  Mingkang Xiang$^{1}$\quad Na Lei$^{1}$\thanks{Corresponding author: nalei@dlut.edu.cn}\quad Shengfa Wang$^{1}$\quad  Zhongxuan Luo$^{1}$\quad Ying He$^{2}$\\
$^{1}$ School of Software, Dalian University of Technology\\
$^{2}$ College of Computing and Data Science, Nanyang Technological University
}

\begin{document}
\maketitle

\begin{figure*}[t] 
  \centering
  \includegraphics[width=\linewidth]{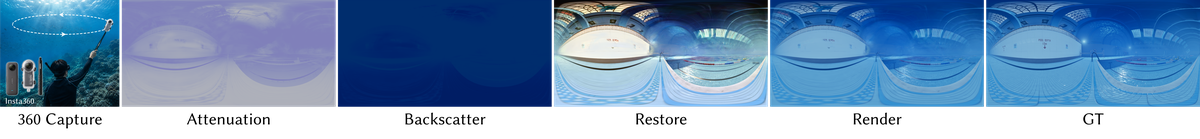}
  \caption{Underwater reconstruction from perspective views is particularly challenging due to limited field of view, severe scattering, and view-dependent degradation, which often lead to incomplete geometry and inconsistent appearance estimation. In contrast, our method leverages omnidirectional observations together with an explicit underwater image formation model to provide stronger global constraints for reconstruction. This enables more consistent 360$^\circ$ scene recovery and higher-quality rendering under challenging underwater conditions.}
  \label{fig:teaser}
\end{figure*}

\begin{abstract}
    Underwater scene reconstruction is essential for immersive exploration of aquatic environments, yet remains challenging due to complex participating-media effects such as absorption and scattering, as well as the limited field of view (FoV) of conventional cameras. Although combining panoramic imaging with 3D Gaussian Splatting (3DGS) offers a promising direction for photorealistic underwater rendering, traditional 3DGS struggles with both spherical projection distortion and underwater medium degradation.
    In this paper, we propose \textbf{Underwater360}, a physics-informed omnidirectional 3DGS framework for underwater panoramic scene reconstruction. First, we introduce an Omnidirectional Gaussian Splatting module that performs ray casting directly in spherical camera space instead of relying on 2D projection approximations, thereby reducing geometric distortions under 360$^\circ$ FoV. Second, we design a physics-based appearance-medium modeling architecture with pose-conditioned appearance embeddings to explicitly decouple intrinsic scene radiance from depth-dependent backscatter and attenuation, enabling physically grounded scene appearance restoration. Finally, we establish a new panoramic underwater benchmark dataset containing both synthetic and real-world scenes. Extensive experiments demonstrate that Underwater360 achieves superior performance in underwater novel view synthesis and scene appearance restoration, delivering improved rendering quality and cross-view consistency in complex underwater environments. The code and datasets are released at \url{https://github.com/SwcK423/Underwater360}
\end{abstract}

\section{Introduction}
\label{sec:introduction}
\begin{figure}[!t]
  \centering
  \setlength{\tabcolsep}{1.0pt}
  \renewcommand{\arraystretch}{1.0}
  \begin{tabular}{ccc}
    \includegraphics[width=0.32\columnwidth]{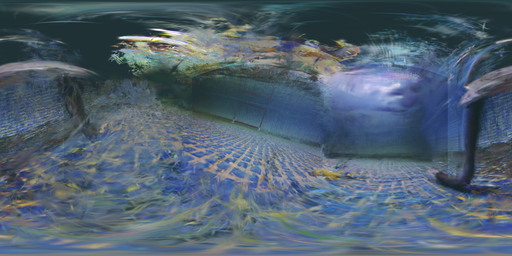} &
    \includegraphics[width=0.32\columnwidth]{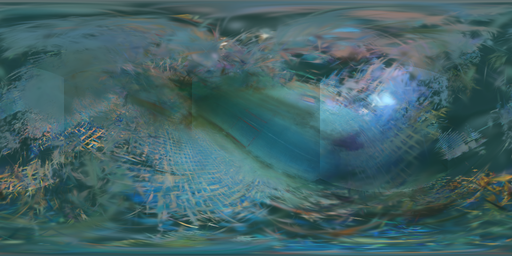} &
    \includegraphics[width=0.32\columnwidth]{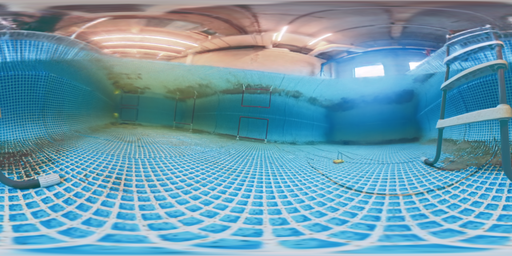}\\
    \small Sparse Input &
    \small Dense Input &
    \small Panorama Input
  \end{tabular}
  \vspace{-0.1in}
  \caption{Challenges of conventional underwater capture with narrow-field cameras. Sparse views are easier to register but provide incomplete scene coverage, whereas dense views improve coverage at the cost of less reliable registration under scattering and attenuation. In contrast, omnidirectional panorama capture offers more complete visual coverage and provides a more suitable basis for robust underwater scene reconstruction.}
  \label{fig:traditional-camera}
  \vspace{-0.1in}
\end{figure}

High-fidelity reconstruction of underwater 3D scenes is essential for immersive multimedia applications. In terrestrial settings, traditional reconstruction is achieved by capturing dense multi-view images with standard perspective cameras. 
However, replicating this paradigm underwater is considerably more challenging. The aquatic environment is often highly unstable, with currents, suspended particles, and dynamic illumination variations, while underwater imaging further suffers from severe optical degradation caused by scattering and absorption~\cite{acuwcv}. These factors substantially weaken cross-view correspondence and often make conventional Structure-from-Motion (SfM) pipelines ~\cite{colmap} less reliable, leading to unstable camera pose estimation and visible reconstruction artifacts. As shown in Fig.~\ref{fig:traditional-camera}, conventional acquisition using standard narrow-field cameras requires a large number of overlapping views to cover a scene, yet reliable reconstruction is still difficult in practice. Sparse views are easier to register, but provide incomplete coverage, whereas dense views improve coverage at the cost of less reliable registration under underwater scattering and attenuation.

In this context, omnidirectional imaging offers a promising alternative for underwater 3D scene reconstruction. By capturing a 360$^\circ$ field of view in a single observation, panoramic cameras provide substantially richer scene coverage and reduce the dependence on densely sampled narrow-field views and carefully designed camera trajectories. 
Such omnidirectional observations can also alleviate temporal inconsistencies across diverse captures, while offering broader contextual cues for geometry-aware reconstruction and immersive rendering.
However, directly applying existing reconstruction and rendering methods to underwater panoramic imagery remains nontrivial, as spherical imaging geometry and participating-media degradation jointly introduce challenges that are not addressed by standard perspective-based pipelines.

Recent advances in differentiable rendering like neural radiance fields (NeRF)~\cite{nerf}, and especially 3D Gaussian Splatting (3DGS)~\cite{3dgs}, have enabled efficient scene representation and high-quality novel view synthesis. In underwater settings, several methods~\cite{seathrunerf, seasplat, uwgs, watersplatting, gaussiansplashing, 3duir} have incorporated underwater image formation models~\cite{uifm} into NeRF- or Gaussian-based frameworks to account for attenuation and backscatter, improving reconstruction fidelity under participating-media degradation. Meanwhile, omnidirectional rendering methods have extended neural scene representations to 360$^\circ$ imagery by adapting projection and sampling strategies to spherical imaging geometry~\cite{egonerf, 360gs, odgs, omnigs, spags}. However, these two lines of work have evolved separately. Existing underwater reconstruction methods are primarily designed for perspective imagery and often rely on implicit neural components or computationally expensive volumetric rendering, which limit efficiency and scalability. In contrast, existing omnidirectional rendering methods mainly target scenes with clear air and do not explicitly model underwater light transport. Overall, current approaches do not adequately address the coupled challenges of spherical projection distortion, medium-induced appearance degradation, and stable geometry recovery in underwater panoramic scenes.

To overcome these limitations, we propose \textit{Underwater360}, a physically geometry-informed omnidirectional 3DGS framework tailored for complex underwater scenes. To the best of our knowledge, this is the first work to utilize panoramic images for underwater 3D scene reconstruction, accompanied by the first benchmark dataset comprising both synthetic and real-world underwater 360$^\circ$ scenes. Rather than adapting existing 2D rasterizers, we introduce an Omnidirectional Gaussian Splatting module that perform spherical ray casting directly in the camera space, thereby effectively mitigating the distortions introduced by 2D projection approximations. To tackle underwater medium degradation, we further design a physics-driven appearance-medium modeling architecture, which disentangles intrinsic scene radiance from distance-dependent attenuation and backscatter through a lightweight network guided by depth priors.

In summary, our main contributions are as follows:
\begin{itemize}
    \item We study underwater omnidirectional scene reconstruction and novel view synthesis from panoramic images, and establish a unified formulation that couples omnidirectional Gaussian rendering with underwater image formation.
    \item We propose a physics-aware omnidirectional Gaussian framework that extends panoramic Gaussian splatting with underwater appearance modeling and explicit participating-medium estimation, enabling joint reconstruction of geometry, view-dependent appearance, and medium degradation.
    \item We offer a new underwater panoramic benchmark with both synthesized and real-world data to support training and evaluation for this community.
\end{itemize}

\begin{figure*}[!t]
    \centering
    \includegraphics[width=0.99\linewidth]{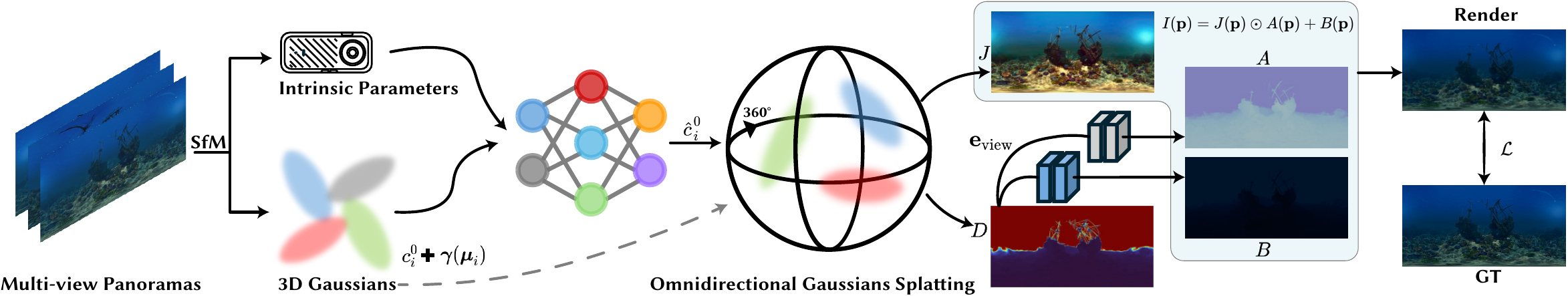}
    \caption{\textbf{Overview of Underwater360}. Our framework combines omnidirectional Gaussian Splatting with underwater image formation modeling for panoramic underwater scene reconstruction. Starting from multi-view panoramas, we estimate camera parameters and initialize 3D Gaussians. A pose-conditioned appearance module adjusts the Gaussians base color using view-dependent embeddings, producing corrected intrinsic appearance for rendering. We then perform omnidirectional Gaussian splatting directly in spherical camera space to obtain the clean scene radiance $J$ and depth $D$. Based on the predicted depth and view embedding $\mathbf{e}_{\text{view}}$, two lightweight convolutional networks estimate the attenuation map $A$ and backscatter component $B$, respectively. Finally, the degraded underwater observation is synthesized according to the image formation model $I(\mathbf{p}) = J(\mathbf{p}) \odot A(\mathbf{p}) + B(\mathbf{p})$ and optimized against the GT panorama using the reconstruction loss $\mathcal{L}$.}
    \vspace{-0.15in}
    \label{fig:pipeline}
\end{figure*}

\section{Related Work}
\label{sec:related-work}

\subsection{Underwater 3D Scene Reconstruction}

Underwater 3D reconstruction is challenging because scene geometry must be recovered jointly with wavelength-dependent light transport in a participating medium~\cite{acuwcv}. Early approaches extend classical underwater image restoration pipelines to multi-view settings, typically by introducing underwater image formation models (UIFM)~\cite{uifm} to estimate attenuation and backscatter. While these physics-based formulations are interpretable, they usually rely on simplified assumptions or strong priors, which makes parameter estimation unstable in complex real scenes~\cite{seathru}.

Recent neural rendering methods alleviate some of these limitations by exploiting multi-view consistency. SeaThru-NeRF~\cite{seathrunerf}, for example, incorporates UIFM into volumetric rendering and explicitly decomposes direct transmission and backscatter. Later methods further improve robustness with additional priors such as illumination decomposition or uncertainty modeling. However, NeRF-style approaches remain computationally expensive due to dense ray sampling and repeated network evaluation, and they scale poorly to large scenes or high-resolution rendering.

More recent work adopts 3D Gaussian Splatting (3DGS) for improved efficiency. Methods such as WaterSplatting~\cite{watersplatting}, SeaSplat~\cite{seasplat}, UW-GS~\cite{uwgs}, and Aquatic-GS~\cite{aquaticgs} combine explicit Gaussian representations with neural medium fields or physics-guided regularization to model underwater degradation. Although these methods substantially accelerate rendering, they often introduce additional implicit parameterizations or auxiliary networks for the medium, which increases optimization complexity. 3D-UIR~\cite{3duir} further separates geometry reconstruction from medium estimation to improve robustness in heavily degraded regions, but it is still designed for perspective imagery and does not address omnidirectional consistency.

In parallel, appearance disentanglement techniques, such as latent embeddings in NeRF-W~\cite{nerfw}, have shown that separating intrinsic scene appearance from observation-dependent factors can improve reconstruction under varying capture conditions. Nevertheless, existing underwater methods still face a trade-off between physical interpretability, optimization stability, and rendering efficiency, and most are designed for perspective imagery rather than omnidirectional input.

\subsection{Omnidirectional Novel View Synthesis}

Omnidirectional novel view synthesis introduces challenges beyond standard perspective reconstruction, including spherical projection distortion, non-uniform sampling, and the need for consistent full-view rendering over 360$^\circ$ imagery. Traditional omnidirectional reconstruction pipelines rely on spherical geometry and multi-view optimization tools such as OpenMVG~\cite{openmvg} to recover sparse or semi-dense structures, but these representations do not directly support photorealistic rendering.

NeRF-based methods extend radiance fields to panoramic scenes through spherical parameterizations or distortion-aware sampling strategies. EgoNeRF~\cite{egonerf} and 360Roam~\cite{360roam} improve reconstruction quality in large-scale 360$^\circ$ environments, but they still inherit the efficiency limitations of volumetric rendering and remain costly for high-resolution panoramic synthesis.
Recent omnidirectional Gaussian splatting methods address this issue by adapting Gaussian projection and rasterization to spherical domains. 360-GS~\cite{360gs}, ODGS~\cite{odgs}, and OmniGS~\cite{omnigs} modify projection or splatting schemes for panoramic images, while SPaGS~\cite{spags} introduces omnidirectional ray--splat intersection and tangent-plane-based bounding strategies for efficient and accurate rendering. Although these methods achieve strong performance in clear scenes, they generally assume standard image formation and do not account for underwater attenuation, backscatter, or appearance degradation.

Our work extends omnidirectional Gaussian splatting to underwater environments by coupling spherical rendering with explicit medium modeling, enabling consistent 360$^\circ$ reconstruction under severe underwater degradation.

\section{Preliminaries}
\label{sec:preliminaries}

\subsection{3DGS}
3DGS~\cite{3dgs} represents a 3D scene using a set of anisotropic 3D Gaussians. Each Gaussian kernel is parameterized by a center position $\mu_i\in\mathbb{R}^3$, a 3D covariance matrix $\Sigma_i\in\mathbb{R}^{3\times 3}$, an opacity value $o_i\in[0,1]$, and a view-dependent color $c_i$ represented by spherical harmonics (SH) coefficients. The spatial distribution of a 3D Gaussian is defined as 
\begin{equation}
    G_i(x) = \exp \left(-\frac{1}{2}(x - \mu_i)^T \Sigma_i^{-1}(x - \mu_i) \right),
\end{equation}
where $\Sigma_i$ is factorized by a scaling matrix $S_i$ and a rotation matrix $R_i$ as $\Sigma=R_iS_iS_i^TR_i^T$. 
For novel view synthesis, 3DGS employs a tile-based rasterizer. The Gaussians are projected onto the 2D image plane, and the final pixel color $C$ is computed using alpha compositing along the depth-sorted Gaussians as
\begin{equation}
    C(x) = \sum_{i=1}^{N} c_i \alpha_i(x) \prod_{j=1}^{i-1} (1 - \alpha_j(x)),
\end{equation}
where $\alpha_i(x)=o_iG_i(x)$ and $N$ denote the number of Gaussians that cover a pixel. While highly efficient, the standard 3DGS rasterization pipeline fundamentally assumes a linear perspective camera model and a clear, non-participating medium, which limits its direct application to underwater panoramic scenes.

\subsection{Underwater Image Formation Model}
\label{sec:uifm}
Underwater imaging is governed by the combined effects of direct signal attenuation and medium backscatter. For the viewing ray associated with pixel $\mathbf{p}=(p_x, p_y)$, UIFM~\cite{uifm} model the observed color $I(\mathbf{p})$ as:
\begin{equation}
I(\mathbf{p}) = J(\mathbf{p}) \odot A(\mathbf{p}) + B(\mathbf{p}),
\label{eq:uifm}
\end{equation}
where $J(\mathbf{p})$ is the scene radiance in the absence of water-induced degradation, $A(\mathbf{p})$ is the transmission map that accounts for attenuation along the ray, and $B(\mathbf{p})$ represents the backscatter component, and $\odot$ denotes element-wise multiplication.
The attenuation term is modeled as an exponential decay with respect to the ray depth $D(\mathbf{p})$:
\begin{equation}
A(\mathbf{p}) = \exp\left(-\beta^D(\mathbf{p}) \cdot D(\mathbf{p})\right),
\end{equation}
where $\beta^D(\mathbf{p})$ denotes the attenuation coefficient for direct component. The backscatter term is:
\begin{equation}
B(\mathbf{p}) = B^\infty \left(1 - \exp(-\beta^B(\mathbf{p})\cdot D(\mathbf{p}))\right) ,
\end{equation}
where $B^\infty$ is the value of the backscatter at infinite distance, $\beta^B(\mathbf{p})$ is the attenuation coefficient for the backscatter component. 

\begin{figure}[!t]
    \centering
    \includegraphics[width=0.9\linewidth]{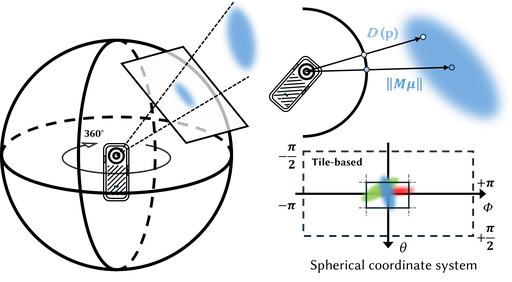}
    \caption{Illustration of Omnidirectional imaging model under ERP projection. A panoramic camera observes the scene over the full viewing sphere, and each Gaussian is intersected by an omnidirectional ray associated with pixel $\mathbf{p}$. The ray-wise depth $D(\mathbf{p})$ and the projected Gaussian footprint are computed in the local tangent plane, after which the contribution is rasterized onto the ERP image using a tile-based scheme.}
    \vspace{-0.15in}
    \label{fig:ERP}
\end{figure}

\section{Method}
\label{sec:method}
\begin{figure*}[!t]
\centering

\begin{tabular}{@{}c@{\,}c@{\,}c@{\,}c@{\,}c@{\,}c@{\,}c@{\,}}
\vspace{1.5pt}
 & 
\small GT & 
\small SeaThru-NeRF &
\small 3DGS &
\small 3D-UIR &
\small SPaGS &
\small Ours \\
\vspace{1.5pt}
\adjustbox{valign=c}{\rotatebox{90}{\scriptsize\text{deepsea}}} &
\includegraphics[width=.16\textwidth,valign=c]{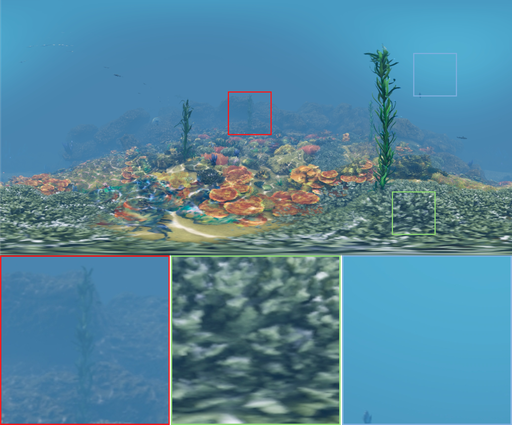} &
\includegraphics[width=.16\textwidth,valign=c]{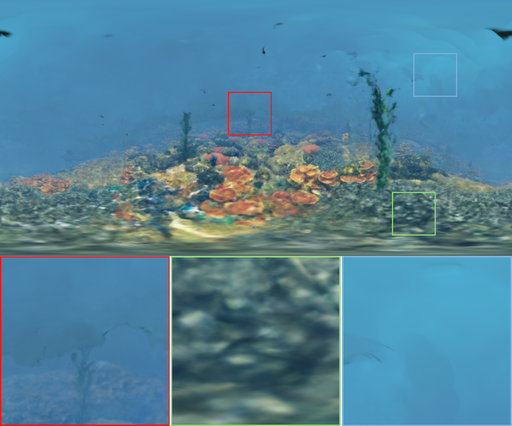} &
\includegraphics[width=.16\textwidth,valign=c]{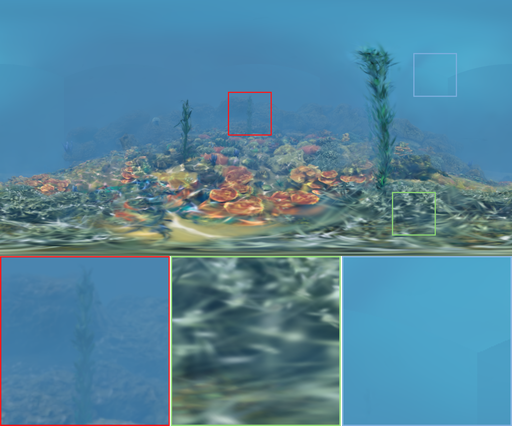} &
\includegraphics[width=.16\textwidth,valign=c]{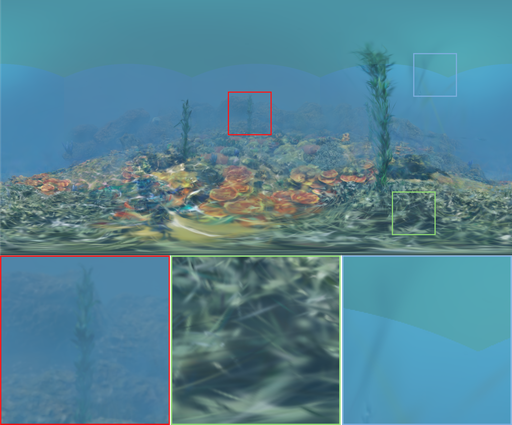} &
\includegraphics[width=.16\textwidth,valign=c]{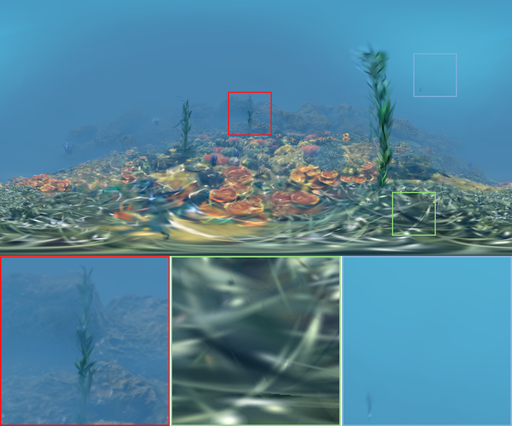} &
\includegraphics[width=.16\textwidth,valign=c]{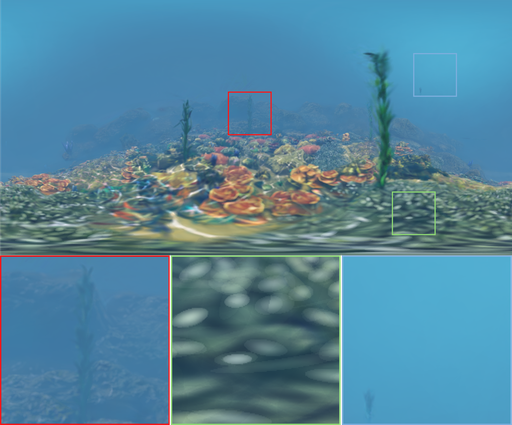} \\
\vspace{1.5pt}
\adjustbox{valign=c}{\rotatebox{90}{\scriptsize\text{rocks}}} &
\includegraphics[width=.16\textwidth,valign=c]{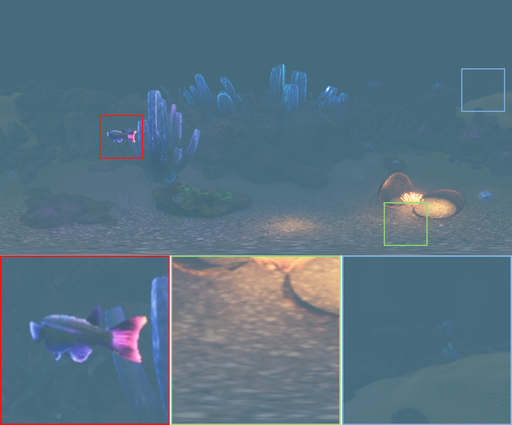} &
\includegraphics[width=.16\textwidth,valign=c]{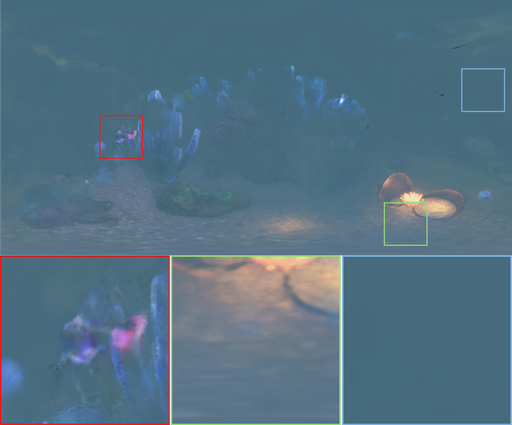} &
\includegraphics[width=.16\textwidth,valign=c]{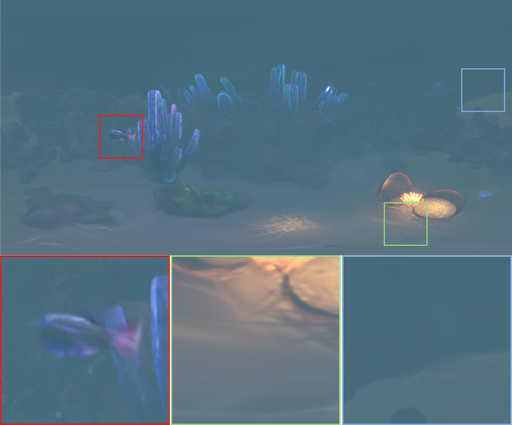} &
\includegraphics[width=.16\textwidth,valign=c]{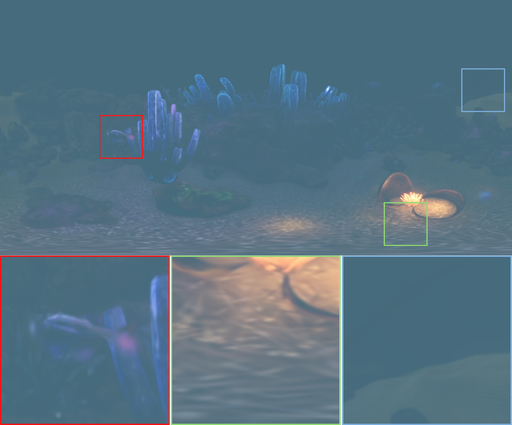} &
\includegraphics[width=.16\textwidth,valign=c]{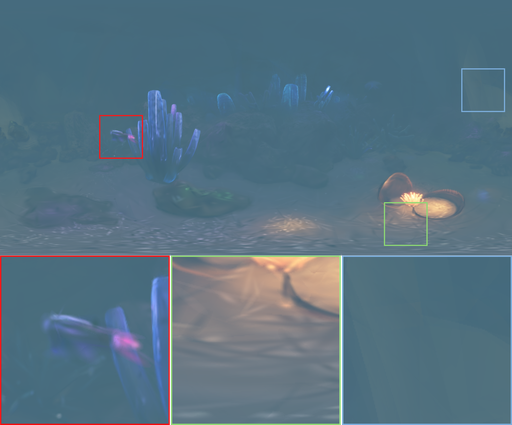} &
\includegraphics[width=.16\textwidth,valign=c]{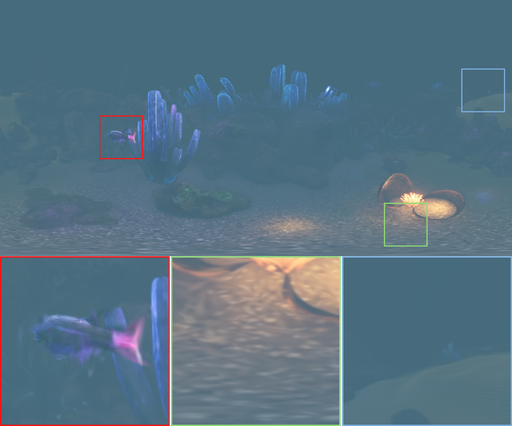} \\
\vspace{1.5pt}
\adjustbox{valign=c}{\rotatebox{90}{\scriptsize\text{ship}}} &
\includegraphics[width=.16\textwidth,valign=c]{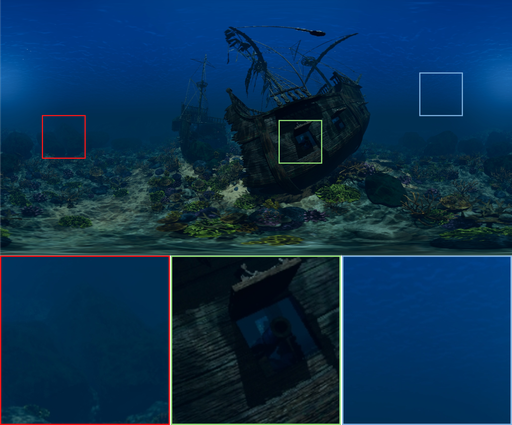} &
\includegraphics[width=.16\textwidth,valign=c]{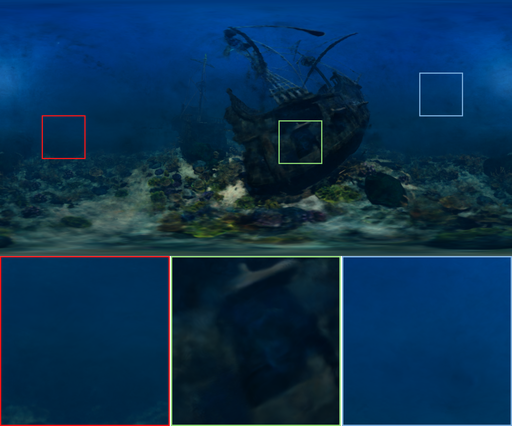} &
\includegraphics[width=.16\textwidth,valign=c]{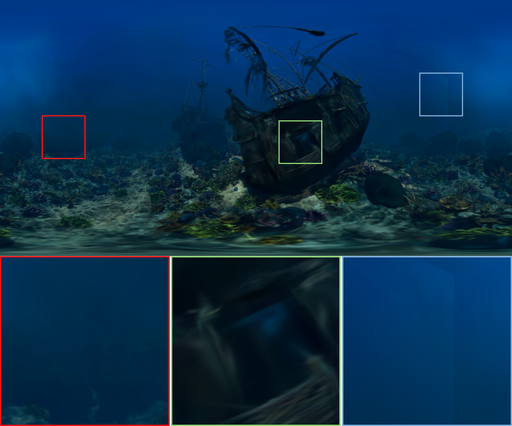} &
\includegraphics[width=.16\textwidth,valign=c]{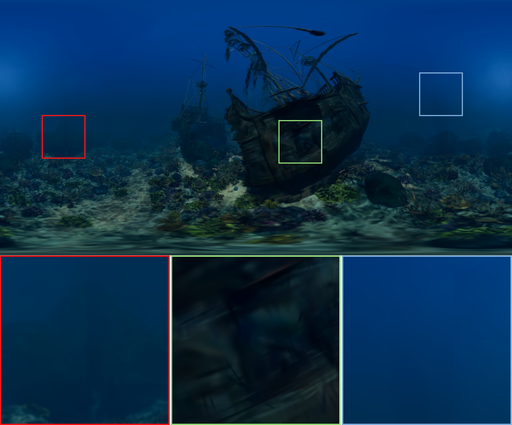} &
\includegraphics[width=.16\textwidth,valign=c]{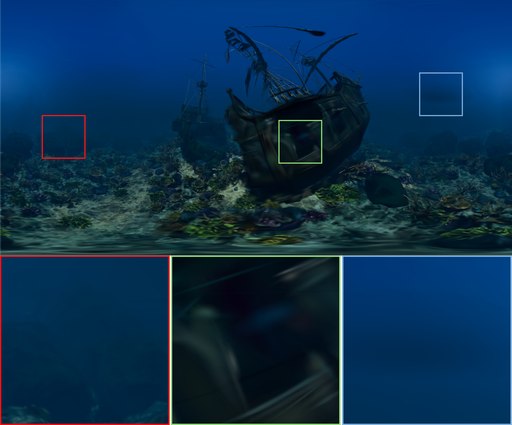} &
\includegraphics[width=.16\textwidth,valign=c]{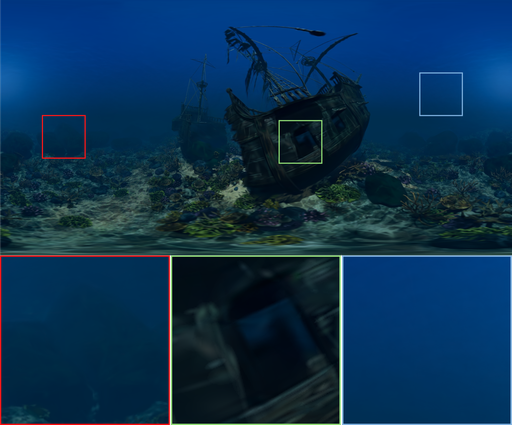} \\
\vspace{1.5pt}
\adjustbox{valign=c}{\rotatebox{90}{\scriptsize\text{scene4}}} &
\includegraphics[width=.16\textwidth,valign=c]{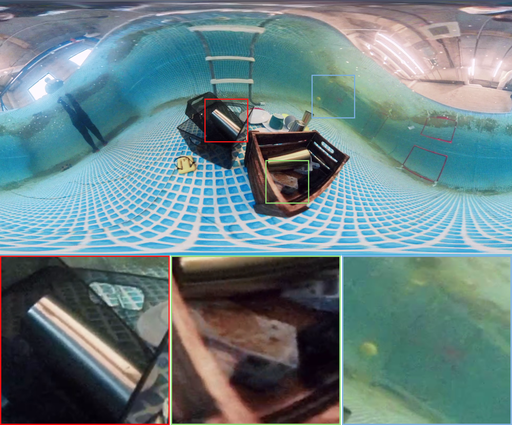} &
\includegraphics[width=.16\textwidth,valign=c]{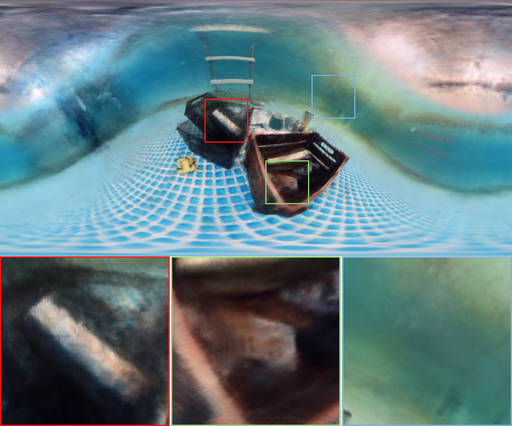} &
\includegraphics[width=.16\textwidth,valign=c]{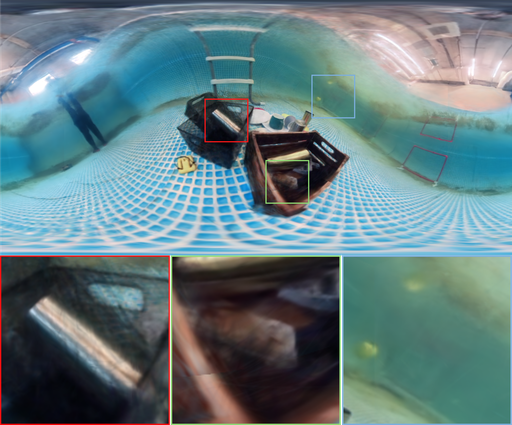} &
\includegraphics[width=.16\textwidth,valign=c]{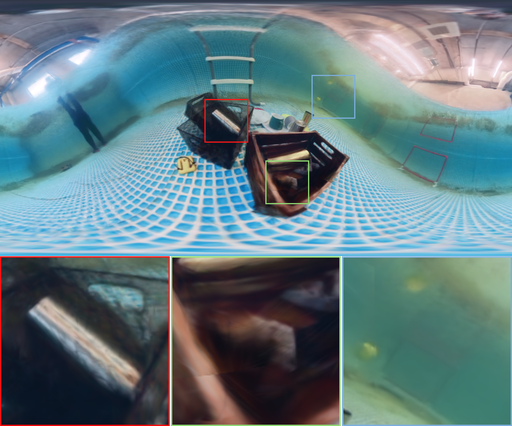} &
\includegraphics[width=.16\textwidth,valign=c]{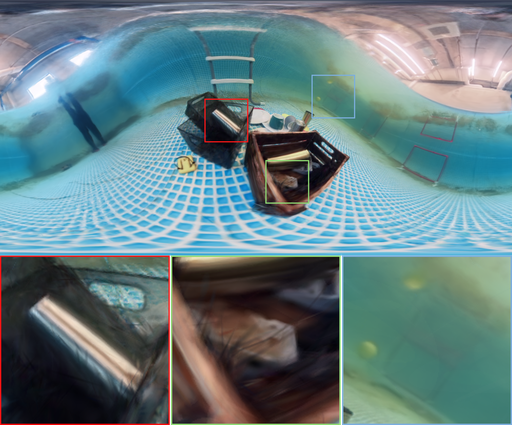} &
\includegraphics[width=.16\textwidth,valign=c]{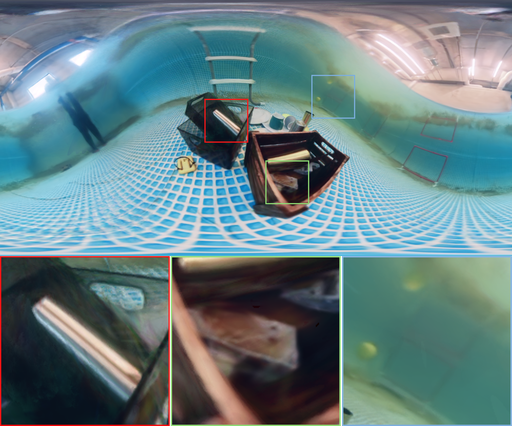} \\

\end{tabular}
\vspace{-0.08in}
\caption{Qualitative comparison on both synthetic and real underwater scenes. The boxed regions are enlarged for detailed comparison. Compared with prior methods, our approach recovers clearer local details and more faithful scene appearance.}
\vspace{-0.15in}
\label{fig:compare}
\end{figure*}
We present a physics-guided framework for underwater panoramic scene reconstruction based on 3DGS. As shown in Fig.~\ref{fig:pipeline}, we explicitly models underwater image formation and jointly recovers scene geometry, intrinsic appearance, and medium properties.

\subsection{Omnidirectional Gaussian Splatting}
\label{subsec:ogs}
Unlike standard cameras, panoramic cameras capture omnidirectional visual observations over a full 360$^\circ$ FoV. These observations are commonly represented using the equirectangular projection (ERP), which maps spherical directions to a 2D image domain, as shown in  Fig.~\ref{fig:ERP}. To avoid the severe geometric distortion caused by projecting 3D Gaussians onto ERP images, we perform Gaussian splatting directly in the spherical camera space. Instead of approximating Gaussian footprints through local affine projection in the 2D image plane, our formulation evaluates Gaussian contributions along omnidirectional viewing rays, yielding ray-splat intersection depth for subsequent underwater modeling.

For an ERP image of width $W$ and height $H$, a pixel coordinate $\mathbf{p}=(p_x, p_y)$ corresponds to spherical coordinates $(\phi, \theta)$, where  $\phi\in[-\pi, \pi]$ denotes longitude and $\theta\in[-\pi/2, \pi/2]$ denotes latitude:
\begin{equation}
    \phi=(\frac{p_x+0.5}{W}-0.5)\cdot 2\pi,\quad \theta=(\frac{p_y+0.5}{H}-0.5)\cdot \pi.
\end{equation}

The corresponding viewing ray direction in camera coordinates is given by
\begin{equation}
    \mathbf{d}(\mathbf{p})=(\sin\phi\cos\theta, \sin\theta, -\cos\phi\cos\theta)^{\top}.
\end{equation}

To evaluate the contribution of each Gaussian under the true omnidirectional geometry, we define two ray-aligned planes in camera space with normals
\begin{equation}
    \mathbf{n}_x=(\cos(\phi), 0, \sin(\phi))^{\top},\quad \mathbf{n}_y=\frac{\mathbf{d}(\mathbf{p})\times \mathbf{n}_x}{\|\mathbf{d}(\mathbf{p})\times \mathbf{n}_x\|},
\end{equation}
and represent them in homogeneous form as $\pi_x=(\mathbf{n}^{\top}_x, 0)^{\top}$, $\pi_y=(\mathbf{n}^{\top}_y, 0)^{\top}$. 
For the $i$-th Gaussian, its local-to-world transform $T_i$ is:
\begin{equation}
    T_i=\left(\begin{array}{cc}
    R_i S_i & \mu_i \\
    \mathbf{0} & 1
    \end{array}\right) .
\end{equation}

Given the world-to-camera transform $M$, the ray constraints in the local splat coordinate system are then obtained by
\begin{equation}
    \pi^s_{x/y, i}=(MT_i)^{\top}\pi_{x/y} .
\end{equation}
These local coordinates are used to compute the Gaussian response $\rho_i^2(\mathbf{p})$ and the corresponding depth ordering value, leading to the opacity contribution
\begin{align}
    \alpha_i(\mathbf{p})&=o_i\exp(-\frac{1}{2}\rho_i^2(\mathbf{p})) , \\
    \rho_i(p)^2&=\left(p-\mu_i\right)^T \Sigma_i^{-1}\left(p-\mu_i\right). \nonumber
\end{align}

After sorting visible Gaussians along each viewing ray, we perform visibility-aware compositing to obtain both rendered color and depth.
In implementation, this omnidirectional splatting module is built upon SPaGS~\cite{spags}, which provides an efficient ray-based rendering framework for spherical panoramas. Since our contribution focuses on underwater appearance-medium modeling rather than panoramic splatting itself, we retain only the core geometric formulation here and refer readers to SPaGS for full derivation and acceleration.

\subsection{Intrinsic Appearance Embedding}
\label{sec:appearance}
Directly assigning RGB attributes to Gaussians is insufficient for underwater scenes, where observed colors are strongly entangled with medium degradation. 
To disentangle scene appearance from water-induced effects, we equip each Gaussian with a learnable color correction model. 
For Gaussian $i$ with diffuse color $c_i^0$, we encode the camera pose as a latent vector $\mathbf{e}_{\text{view}}$ through a lightweight MLP (two layers with 128 hidden units) and compute spatial Fourier features $\boldsymbol{\gamma}(\mu_i)$. Then we use a lightweight MLP $\mathcal{F}$ (two layers with 128 hidden units) to predict the affine correction parameters
\begin{equation}
(\boldsymbol{\beta}_i, \boldsymbol{\eta}_i) = \mathcal{F}(c_i^0, \mathbf{e}_{\text{view}}, \boldsymbol{\gamma}(\mu_i)),
\end{equation}
where $\boldsymbol{\eta}_i\in\mathbb{R}^3$ and $\boldsymbol{\beta}_i\in\mathbb{R}^3$ are the channel-wise scale and shift factors, respectively. The corrected diffuse color is computed as
\begin{equation}
\hat{c}_i^0 = \boldsymbol{\eta}_i \odot c_i^0 + \boldsymbol{\beta}_i.
\end{equation}

We use $\hat{c}_i^0$ in place of the original diffuse component when evaluating Gaussian radiance, while keeping the higher-order spherical harmonics coefficients to preserve view-dependent effects. The corrected radiance is then composited through the omnidirectional splatting module in Sec.~\ref{subsec:ogs} to produce the clean appearance $J$ input for subsequent underwater image formation. In this way, the base Gaussian color is prevented from directly absorbing medium-related appearance shifts, leading to a more stable and physically compatible scene representation.

\subsection{Scattering Medium Modeling}
\label{sec:scattering}

To simulate underwater image formation from the rendered depth map $D$, we employ two lightweight neural networks to estimate the backscatter and attenuation terms, respectively, under the physical model in Eq.~\eqref{eq:uifm}. Specifically, the rendered scene appearance from Sec.~\ref{sec:appearance} is treated as the clean radiance, while the ray-wise depth serves as the geometric cue for modeling distance-dependent underwater degradation.

\textbf{Backscatter estimation.}
The backscatter component arises from ambient light scattered into the line of sight by suspended particles, and gradually approaches a saturated waterlight color as the propagation distance increases. To model this effect, we use a Backscatter-Net that takes the single-channel depth map $D\in\mathbb{R}^{1\times H\times W}$ as input and predicts two coefficients through parallel $1\times1$ convolutional layers:
\begin{equation}
\beta_{b/r} = \operatorname{Softplus}(\operatorname{Conv}_{b/r}(D)), 
\label{eq:bs_conv}
\end{equation}
where both $\beta_b, \beta_r\in\mathbb{R}^{3\times H\times W}$. Together with two globally learnable color vectors, namely the infinite-distance waterlight $B^\infty\in\mathbb{R}^{3}$ and a residual color $B^{res}\in\mathbb{R}^{3}$, the backscatter image is computed as 
\begin{equation}
B(\mathbf{p}) =
B^\infty\odot \left(1 - e^{-\beta_{b}(\mathbf{p})}\right)
+
B^{res} \odot e^{-\beta_{r}(\mathbf{p})}.
\label{eq:backscatter}
\end{equation}

Here, the first term models the asymptotic saturation of backscatter with increasing distance, while the second residual term captures near-field deviations from the ideal saturation curve.

\textbf{Direct attenuation estimation.}
The attenuation of the direct signal is governed by the propagation distance and the medium-dependent attenuation coefficient. To account for spatially varying attenuation in non-homogeneous water, we employ a DeattenuateNet that takes as input the concatenation of the depth map $D$ and the view embedding $\mathbf{e}_{\text{view}}$. A $1\times1$ convolutional layer predicts $P$ candidate attenuation components:
\begin{equation}
\mathbf{a} = \operatorname{Sigmoid}\bigl( \operatorname{Conv}_a(\; D \oplus \mathbf{e}_{\text{view}} \;) \bigr) \;\in\; \mathbb{R}^{P \times H\times W},
\label{eq:attn_conv}
\end{equation}
where $\oplus$ denotes channel-wise concatenation with appropriate broadcasting. These candidates are fused by learnable weights $\{\lambda_k\in\mathbb{R}\}_{k=1}^P$ to produce the final attenuation coefficient map:
\begin{equation}
\beta^D(\mathbf{p}) = \sum_{k=1}^{P} \lambda_{k} \cdot a_{k}(\mathbf{p}).
\label{eq:beta_d}
\end{equation}

The attenuation factor is then computed as
\begin{equation}
A(\mathbf{p}) = \operatorname{Sigmoid}\left( \exp\left(-\beta^D(\mathbf{p}) \cdot D(\mathbf{p})\right) \right).
\label{eq:attenuation}
\end{equation}

This design yields a bounded attenuation map that remains numerically stable during training, while still preserving the desired distance-decay behavior. 

The estimated backscatter and attenuation fields are jointly optimized with the Gaussian scene representation.

\subsection{Optimization}
\label{sec:inverse}
With the forward underwater image formation model, reconstruction is formulated as a joint inverse problem over the Gaussian scene representation and the learnable parameters of the appearance and degradation modules. Let $\mathcal{G}_\Psi$ denote the set of Gaussian attributes to be optimized, and let $\Theta$ collect the parameters of the color correction, backscatter, and attenuation networks. Given a set of raw underwater observations $\{I^{\text{raw}}_k\}$, we optimize the rendered predictions $\hat{I}_k(\mathcal{G}_\Psi, \Theta)$ by minimizing
\begin{equation}
\min_{\mathcal{G}_\Psi, \Theta} \sum_k
\mathcal{L}\bigl(
\hat{I}_k(\mathcal{G}_\Psi, \Theta),
I_k^{\text{raw}}
\bigr).
\end{equation}

The main supervision is imposed on the reconstructed underwater image using a combination of pixel-wise $L_1$ loss and structural similarity loss
\begin{equation}
\mathcal{L}=\left(1-\lambda_1\right) \mathcal{L}_1+\lambda_1 \mathcal{L}_{\mathrm{D}-\mathrm{SSIM}},
\end{equation}
where $\mathcal{L}_1$ measures absolute color differences and $\mathcal{L}_{\mathrm{D}-\mathrm{SSIM}}$ encourages structural consistency between the rendered image and the raw underwater observation. 

All parameters, $\mathcal{G}_\Psi$ and $\Theta$, are optimized jointly using Adam. Following the adaptive density control strategy of Gaussian splatting, we periodically perform gradient-based splitting and cloning of Gaussians, prune low-opacity primitives, and remove degenerate Gaussians to maintain an efficient and expressive scene representation. To further stabilize rendering across scales and suppress aliasing artifacts, we incorporate a 3D Gaussian filtering strategy inspired by Mip-Splatting~\cite{mipsplatting} during optimization.

\setlength\tabcolsep{0.5em}
\begin{table*}[!htbp]
\vspace{-2pt}
\centering
\caption{Quantitative comparison on the synthetic OmniUW dataset. The methods are grouped into four categories: (1) standard reconstruction without underwater physical modeling, (2) standard reconstruction with underwater physical modeling, (3) omnidirectional reconstruction without underwater physical modeling, and (4) our omnidirectional reconstruction with explicit underwater physical modeling. Our method achieves the best overall performance across most scenes and metrics, demonstrating the benefit of jointly leveraging panoramic observations and underwater image formation priors.}
\vspace{-0.15in}
\label{tab:quantity1}

\resizebox{.99\textwidth}{!}{

\begin{tabular}{@{}l|ccc|ccc|ccc|ccc|ccc}
\midrule
    & \multicolumn{3}{c|}{building} & \multicolumn{3}{c|}{deepsea}  & \multicolumn{3}{c|}{pool} & \multicolumn{3}{c|}{rocks} & \multicolumn{3}{c}{ship} \\ 
\cmidrule(lr){2-4} \cmidrule(lr){5-7} \cmidrule(lr){8-10} \cmidrule(lr){11-13} \cmidrule(lr){14-16} 
    Method & PSNR~$\uparrow$ & SSIM~$\uparrow$ & LPIPS~$\downarrow$ & PSNR~$\uparrow$ & SSIM~$\uparrow$ & LPIPS~$\downarrow$ & PSNR~$\uparrow$ & SSIM~$\uparrow$ & LPIPS~$\downarrow$ & PSNR~$\uparrow$ & SSIM~$\uparrow$ & LPIPS~$\downarrow$ & PSNR~$\uparrow$ & SSIM~$\uparrow$ & LPIPS~$\downarrow$ \\
\midrule
    INGP & \tbest 33.71 & 0.883 & 0.291 & 22.74 & 0.752 & 0.296 & 22.53 & 0.778 & 0.452 & 31.55 & 0.930 & 0.237 & 30.58 & 0.835 & 0.305 \\
    ZipNeRF & 26.57 & 0.827 & 0.385 & 24.28 & 0.771 & 0.310 & 20.80 & 0.782 & 0.413 & 29.04 & 0.933 & 0.259 & 32.52 & 0.870 & 0.259 \\
    EgoNeRF & 33.13 & 0.881 & 0.379 & 23.90 & 0.773 & 0.365 & 25.77 & 0.840 & 0.422 & 33.36 & 0.941 & 0.276 & 31.05 & 0.838 & 0.380 \\
    3DGS & 33.48 & \tbest 0.906 & \tbest 0.263 & \tbest 26.56 & \tbest 0.824 & \tbest 0.241 & 26.25 & 0.878 & 0.258 & \sbest 35.93 & \tbest 0.953 & 0.190 & 32.18 & \tbest 0.875 & 0.259 \\
    2DGS & 33.15 & 0.899 & 0.298 & 26.50 & 0.818 & 0.283 & 26.22 & 0.872 & 0.294 & 34.44 & 0.949 & 0.218 & 31.81 & 0.867 & 0.286 \\
\midrule
    SeaThru-NeRF & 29.61 & 0.879 & 0.314 & 26.01 & 0.798 & 0.269 & 24.69 & 0.840 & 0.337 & 27.23 & 0.926 & 0.261 & 29.20 & 0.842 & 0.310 \\
    SeaSplat & 30.32 & 0.876 & 0.331 & 22.08 & 0.789 & 0.350 & 21.97 & 0.817 & 0.336 & 27.79 & 0.936 & 0.265 & 26.58 & 0.854 & 0.316 \\
    Water-Splatting & 15.84 & 0.673 & 0.394 & 11.81 & 0.567 & 0.419 & 11.86 & 0.575 & 0.635 & 33.19 & 0.935 & 0.200 & 20.12 & 0.769 & 0.367 \\
    UW-GS & 28.18 & 0.860 & 0.323 & 19.80 & 0.759 & 0.378 & 19.63 & 0.771 & 0.408 & 33.18 & 0.949 & 0.203 & 30.57 & 0.866 & 0.265 \\
    Gaussian Splashing & 29.56 & 0.870 & 0.326 & 23.48 & 0.787 & 0.310 & 23.63 & 0.831 & 0.343 & 32.12 & 0.943 & 0.219 & OOM & OOM & OOM \\
    3D-UIR & 31.59 & 0.851 & 0.289 & \sbest 26.60 & \best 0.827 & \best 0.229 & 26.47 & \tbest 0.879 & \sbest 0.240 & \tbest 35.09 & \sbest 0.955 & \tbest 0.167 & \tbest 32.84 & \tbest 0.875 & \tbest 0.253 \\
\midrule
    360-GS & 31.50 & 0.879 & 0.329 & 23.17 & 0.772 & 0.351 & 24.68 & 0.859 & 0.301 & 33.37 & 0.942 & 0.224 & 29.60 & 0.836 & 0.338 \\
    ODGS & 28.58 & 0.853 & \sbest 0.244 & 23.93 & 0.770 & 0.243 & 23.80 & 0.817 & 0.314 & 32.15 & 0.937 & \best 0.139 & 30.02 & 0.830 & \best 0.211 \\
    OmniGS & 30.55 & 0.858 & 0.349 & 23.36 & 0.767 & 0.341 & \tbest 26.79 & 0.875 & 0.305 & 32.52 & 0.943 & 0.221 & 30.54 & 0.836 & 0.360 \\
    SPaGS & \sbest 35.88 & \sbest 0.919 & 0.313 & 26.08 & 0.822 & 0.288 & \best 28.07 & \best 0.908 & \tbest 0.250 & 29.74 & 0.908 & 0.303 & \sbest 33.30 & \sbest 0.889 & 0.302 \\
\midrule
    Ours & \best 36.06 & \best 0.922 & \best 0.243 & \best 27.01 & \sbest 0.825 & \sbest 0.239 & \sbest 27.10 & \sbest 0.901 & \best 0.202 & \best 37.92 & \best 0.964 & \sbest 0.145 & \best 34.19 & \best 0.900 & \sbest 0.219 \\
\midrule
\end{tabular}

}

\vspace{-2pt}
\end{table*}

\setlength\tabcolsep{0.5em}
\begin{table*}[!htbp]
\vspace{-2pt}
\centering
\caption{Quantitative results on the real-world Insta360 dataset. Our method remains consistently strong across scenes, validating its robustness and generalization on real omnidirectional underwater data.}
\vspace{-0.15in}
\label{tab:quantity2}

\resizebox{.99\textwidth}{!}{

\begin{tabular}{@{}l|ccc|ccc|ccc|ccc|ccc}
\midrule
    & \multicolumn{3}{c|}{scene0} & \multicolumn{3}{c|}{scene1}  & \multicolumn{3}{c|}{scene2} & \multicolumn{3}{c|}{scene3} & \multicolumn{3}{c}{scene4} \\ 
\cmidrule(lr){2-4} \cmidrule(lr){5-7} \cmidrule(lr){8-10} \cmidrule(lr){11-13} \cmidrule(lr){14-16} 
    Method & PSNR~$\uparrow$ & SSIM~$\uparrow$ & LPIPS~$\downarrow$ & PSNR~$\uparrow$ & SSIM~$\uparrow$ & LPIPS~$\downarrow$ & PSNR~$\uparrow$ & SSIM~$\uparrow$ & LPIPS~$\downarrow$ & PSNR~$\uparrow$ & SSIM~$\uparrow$ & LPIPS~$\downarrow$ & PSNR~$\uparrow$ & SSIM~$\uparrow$ & LPIPS~$\downarrow$ \\
\midrule
    INGP & 24.38 & 0.678 & 0.340 & 22.84 & 0.654 & 0.399 & 21.92 & 0.644 & 0.430 & 23.05 & 0.693 & 0.395 & 22.07 & 0.674 & 0.404 \\
    ZipNeRF & 20.19 & 0.559 & 0.508 & 16.85 & 0.537 & 0.556 & 21.12 & 0.671 & 0.392 & 17.57 & 0.601 & 0.532 & 16.26 & 0.573 & 0.557 \\
    EgoNeRF & 23.15 & 0.641 & 0.423 & 21.27 & 0.640 & 0.449 & 20.22 & 0.638 & 0.483 & 22.11 & 0.684 & 0.453 & 21.27 & 0.670 & 0.451 \\
    3DGS & 25.08 & 0.739 & 0.285 & 22.72 & 0.671 & 0.357 & \tbest 22.92 & 0.688 & 0.359 & 23.50 & 0.736 & 0.345 & 23.14 & 0.718 & 0.349 \\
    2DGS & 24.52 & 0.735 & 0.306 & 22.59 & 0.683 & 0.366 & 22.63 & \best 0.693 & 0.375 & 22.83 & 0.726 & 0.372 & 22.99 & \best 0.722 & 0.364 \\
\midrule
    SeaThru-NeRF & 23.05 & 0.615 & 0.426 & 16.60 & 0.522 & 0.549 & 19.58 & 0.604 & 0.474 & 20.85 & 0.634 & 0.476 & 20.34 & 0.625 & 0.460 \\
    SeaSplat & 23.92 & 0.718 & 0.312 & 22.72 & 0.679 & 0.364 & 22.36 & 0.684 & 0.381 & 22.64 & 0.716 & 0.375 & 22.58 & \tbest 0.709 & 0.368 \\
    Water-Splatting & 13.87 & 0.470 & 0.516 & 17.64 & 0.577 & 0.456 & 17.85 & 0.587 & 0.469 & 16.02 & 0.578 & 0.468 & 17.05 & 0.591 & 0.468 \\
    UW-GS & 23.71 & 0.696 & 0.330 & 22.32 & 0.654 & 0.381 & 20.86 & 0.629 & 0.433 & 21.58 & 0.674 & 0.405 & 21.85 & 0.684 & 0.392 \\
    3D-UIR & \tbest 25.10 & \tbest 0.745 & \tbest 0.276 & \tbest 23.27 & \tbest 0.690 & \tbest 0.336 & 22.90 & 0.687 & \tbest 0.353 & \tbest 23.52 & \tbest 0.737 & \tbest 0.343 & \tbest 23.26 & \best 0.722 & \tbest 0.343 \\
\midrule
    360-GS & OOM & OOM & OOM & 17.86 & 0.577 & 0.489 & 20.87 & 0.672 & 0.402 & 21.01 & 0.696 & 0.406 & 20.57 & 0.672 & 0.425 \\
    ODGS & OOM & OOM & OOM & 21.93 & 0.638 & \best 0.298 & 21.68 & 0.641 & \best 0.325 & 21.47 & 0.660 & \sbest 0.335 & 21.72 & 0.665 & \sbest 0.326 \\
    OmniGS & 24.97 & 0.730 & 0.325 & \sbest 23.32 & \sbest 0.693 & 0.351 & \best 23.05 & \sbest 0.691 & 0.370 & \sbest 23.62 & 0.734 & 0.359 & \sbest 23.31 & \sbest 0.718 & 0.359 \\
    SPaGS & \sbest 26.03 & \sbest 0.779 & \sbest 0.271 & 22.33 & \best 0.695 & 0.356 & 21.63 & \sbest 0.691 & 0.375 & 23.32 & \sbest 0.745 & 0.360 & 22.46 & \best 0.722 & 0.360 \\
\midrule
    Ours & \best 26.40 & \best 0.785 & \best 0.241 & \best 23.46 & \tbest 0.690 & \sbest 0.310 & \sbest 22.95 & \tbest 0.689 & \sbest 0.329 & \best 24.11 & \best 0.746 & \best 0.313 & \best 23.43 & \sbest 0.718 & \best 0.318 \\
\midrule
\end{tabular}

}

\vspace{-2pt}
\end{table*}

\begin{figure}[!]
\centering

\begin{tabular}{@{}c@{\,}c@{\,}c@{\,}c@{\,}}
\vspace{1.5pt}
 & 
\small Attenuation & 
\small Backscatter &
\small Restore \\
\vspace{1.5pt}
\adjustbox{valign=c}{\rotatebox{90}{\scriptsize\text{SeaThru-NeRF}}} &
\includegraphics[width=.15\textwidth,valign=c]{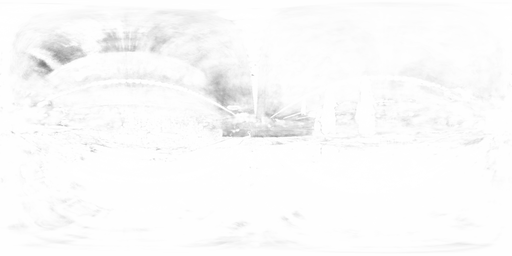} &
\includegraphics[width=.15\textwidth,valign=c]{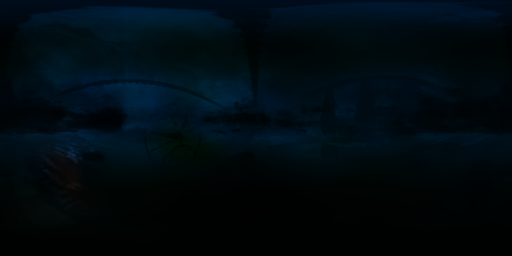} &
\includegraphics[width=.15\textwidth,valign=c]{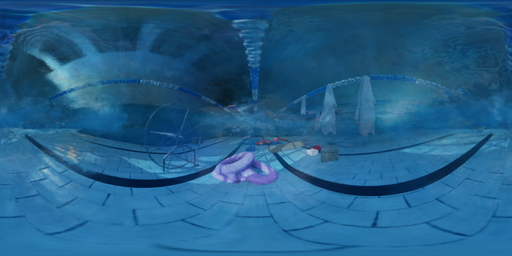} \\
\vspace{1.5pt}
\adjustbox{valign=c}{\rotatebox{90}{\scriptsize\text{SeaSplat}}} &
\includegraphics[width=.15\textwidth,valign=c]{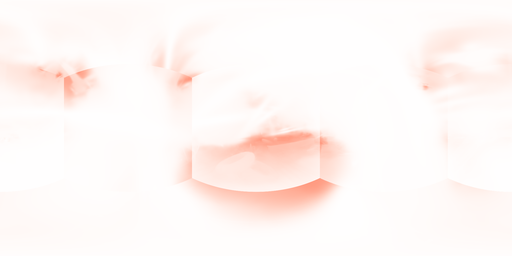} &
\includegraphics[width=.15\textwidth,valign=c]{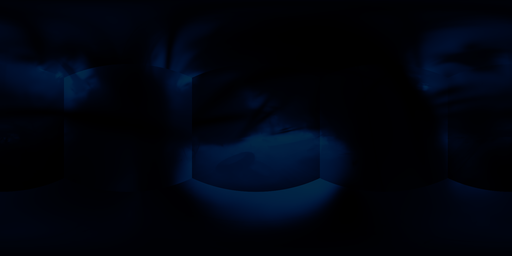} &
\includegraphics[width=.15\textwidth,valign=c]{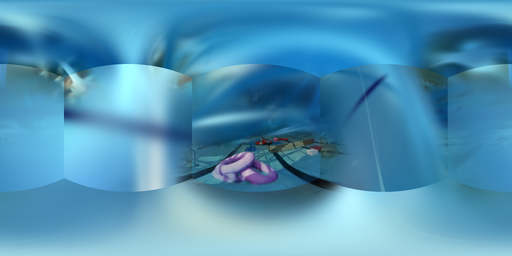} \\
\vspace{1.5pt}
\adjustbox{valign=c}{\rotatebox{90}{\scriptsize\text{3D-UIR}}} &
\includegraphics[width=.15\textwidth,valign=c]{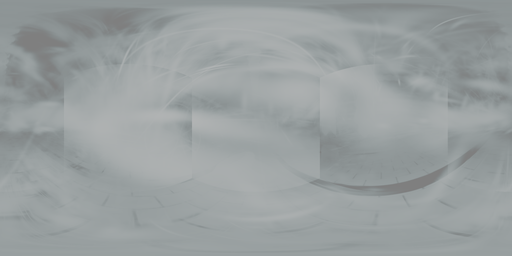} &
\includegraphics[width=.15\textwidth,valign=c]{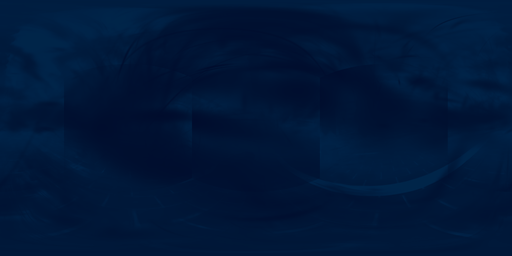} &
\includegraphics[width=.15\textwidth,valign=c]{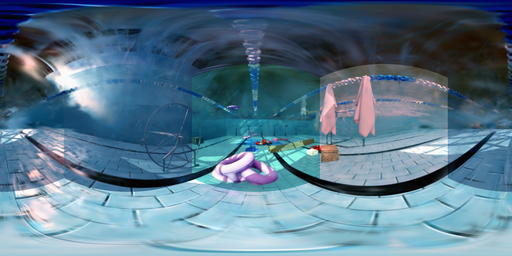} \\
\vspace{1.5pt}
\adjustbox{valign=c}{\rotatebox{90}{\scriptsize\text{Ours}}} &
\includegraphics[width=.15\textwidth,valign=c]{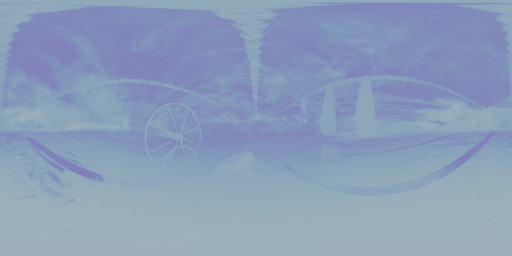} &
\includegraphics[width=.15\textwidth,valign=c]{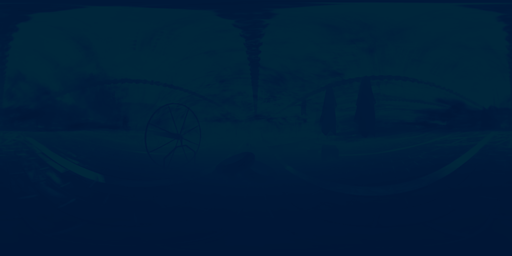} &
\includegraphics[width=.15\textwidth,valign=c]{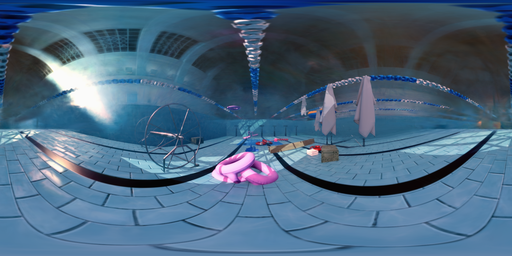} \\

\end{tabular}
\vskip-1ex
\caption{Qualitative comparison of UIFM decomposition on the \textit{pool} scene. From left to right, we show the estimated attenuation maps, backscatter maps, and restored scene appearance. Compared with prior methods, our approach yields more spatially coherent attenuation and backscatter estimation, while recovering a clearer and more faithful scene appearance, demonstrating more effective disentanglement of intrinsic radiance from underwater degradation effects.}
\label{fig:compare_uifm}
\vspace{-0.18in}  
\end{figure}

\section{Experiments}
\label{sec:results}

\subsection{Settings}

\textbf{Datasets.} Since most existing underwater image datasets are captured with standard cameras, we construct two omnidirectional benchmarks for evaluation. Specifically, we build five underwater scenes in Unreal Engine 5 and render omnidirectional images for training and evaluation. The synthetic dataset, \textit{OmniUW}, includes five representative scenes: \textit{building}, \textit{deepsea}, \textit{pool}, \textit{rocks}, and \textit{ship}. Each scene contains 121--149 images at a resolution of 2048 $\times$ 1024. In addition to the synthetic dataset, we collect a real-world underwater panoramic dataset in a swimming pool using an Insta360 X4 camera. We capture five real scenes (scene0--4) with different arrangements of underwater objects made of different materials, forming our real dataset, \textit{Insta360}. Each scene contains 106--209 images at a resolution of 1920 $\times$ 960. For all datasets, we adopt a 1:1 train/test split for evaluation.

\textbf{Baselines.} We compare our method against both NeRF-based methods~\cite{instantngp, zipnerf, egonerf, seathrunerf} and GS-based methods~\cite{3dgs, 2dgs, seasplat, watersplatting, uwgs, gaussiansplashing, 3duir, 360gs, odgs, omnigs, spags}, including approaches designed for standard and omnidirectional cameras. Some baselines~\cite{seathrunerf, seasplat, watersplatting, uwgs, gaussiansplashing, 3duir} also incorporate underwater image formation modeling.
For methods designed for standard cameras, we first decompose each omnidirectional image into six cubemap faces as input. The rendered outputs are then projected back and stitched into omnidirectional images for comparison. We evaluate reconstruction quality using PSNR, SSIM~\cite{ssim} and LPIPS~\cite{lpips}.

\textbf{Implementation.} Our method is implemented in PyTorch and optimized with Adam. Unless otherwise specified, we follow the default training settings of SPaGS, with minor modifications for our underwater modeling pipeline. In particular, we set \textit{densify \_grad \_threshold} to 2e-5.
For perspective-camera baselines, camera poses are estimated using COLMAP. For omnidirectional datasets, we use Agisoft Metashape~\cite{agisoft_metashape} for camera calibration.
All experiments are conducted on a single NVIDIA RTX 5090 or NVIDIA RTX L20 GPU.

\subsection{Comparisons}

\textbf{Qualitative.}
Fig.~\ref{fig:compare} presents qualitative comparisons. Compared with perspective-based baselines, our method yields more consistent panoramic reconstruction, especially for attenuation and backscatter estimation. Compared with panoramic baselines, it better preserves scene details under strong degradation and exhibit fewer floating artifacts and less blur. As illustrated in Fig.~\ref{fig:compare_uifm}, the recovered decomposition of clean radiance, attenuation, and backscatter is also more physically plausible, as it better follows the depth-dependent degradation predicted by underwater image formation.

\noindent\textbf{Quantitative.} 
Tables \ref{tab:quantity1} and \ref{tab:quantity2} summarize quantitative results on the synthetic OmniUW dataset and the real-world Insta360 dataset. Our method achieves the best or second-best performance in most cases, showing that robust underwater reconstruction benefits from jointly leveraging omnidirectional observations and explicit underwater physical modeling. 
The comparisons also reveal that omnidirectional Gaussian splatting is already effective for panoramic underwater reconstruction when the observation are relatively clear. However, under more severe degradation, their performance becomes less stable, indication that explicit underwater image formation modeling is crucial for handling attenuation and backscatter.

\section{Conclusion}
\label{sec:conclusion}

We presented \textit{Underwater360}, a unified framework for physics-aware omnidirectional underwater scene reconstruction that combines omnidirectional Gaussian Splatting with a learnable underwater image formation model. By disentangling scene appearance from water-induced degradation, our method jointly reconstructs the 3D scene while estimating per-pixel attenuation and backscatter directly from input panoramas. To support this task, we further introduce two underwater omnidirectional datasets, \textit{OmniUW} and \textit{Insta360}. Extensive experiments on both synthetic and real data demonstrate the effectiveness of our approach.  More broadly, we hope this work offers a new perspective on underwater scene reconstruction by unifying panoramic scene representation and explicit underwater physical modeling.

Our method still has several limitations. First, like most existing underwater reconstruction methods, it assumes a static scene and does not explicitly model dynamic elements such as moving marine life, suspended particles, or temporal appearance fluctuations. Handling real underwater video in the presence of such transient effects requires a more flexible temporal reconstruction mechanism. Second, our omnidirectional camera model does not account for refraction introduced by the camera housing, which may lead to systematic geometric bias, especially in close-range scenes. Third, in extremely turbid or texture-sparse environments, where depth cues are weak and ambiguous, the estimation of medium parameters may become unstable. This suggests the need for stronger physical priors or additional sensing modalities such as depth measurements.

Future work will extend the framework in three directions: dynamic underwater scene modeling through temporal appearance embeddings or deformation fields, refraction-aware geometry for imaging through underwater housings and planar ports, and more compact scene representations via tensor decomposition for deployment on resource-constrained platforms such as autonomous underwater vehicles. The source code and datasets will be released publicly to support future research in underwater 3D vision. We hope \textit{Underwater360} can serve as a promising step toward a new solution paradigm for robust underwater scene reconstruction and omnidirectional underwater 3D vision.

\newpage
{
    \small
    \bibliographystyle{ieeenat_fullname}
    \bibliography{ref}
}

\begin{figure*}[!htbp]
\centering

\begin{tabular}{@{}c@{\,}c@{\,}c@{\,}c@{\,}c@{\,}c@{\,}c@{\,}}
\vspace{1.5pt}
 & 
\small building & 
\small deepsea &
\small pool &
\small rocks &
\small ship &
\small scene4 \\
\vspace{1.5pt}
\adjustbox{valign=c}{\rotatebox{90}{\scriptsize\text{GT}}} &
\includegraphics[width=.16\textwidth,valign=c]{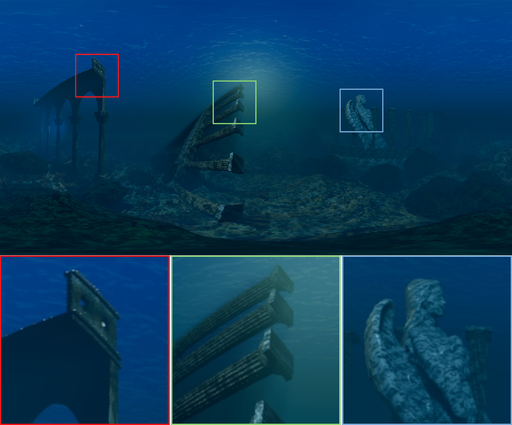} &
\includegraphics[width=.16\textwidth,valign=c]{figs/compare/deepsea_gt.png} &
\includegraphics[width=.16\textwidth,valign=c]{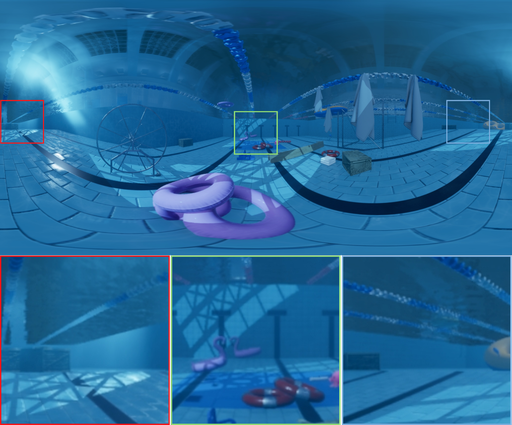} &
\includegraphics[width=.16\textwidth,valign=c]{figs/compare/rocks_gt.png} &
\includegraphics[width=.16\textwidth,valign=c]{figs/compare/ship_gt.png} &
\includegraphics[width=.16\textwidth,valign=c]{figs/compare/scene4_gt.png} \\
\vspace{1.5pt}
\adjustbox{valign=c}{\rotatebox{90}{\scriptsize\text{INGP}}} &
\includegraphics[width=.16\textwidth,valign=c]{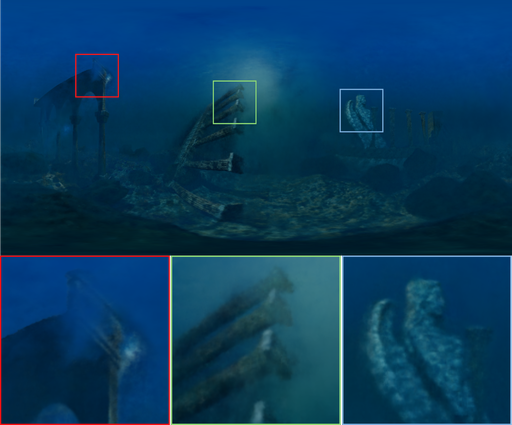} &
\includegraphics[width=.16\textwidth,valign=c]{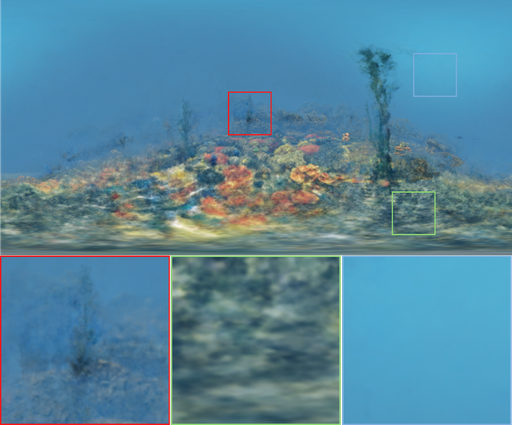} &
\includegraphics[width=.16\textwidth,valign=c]{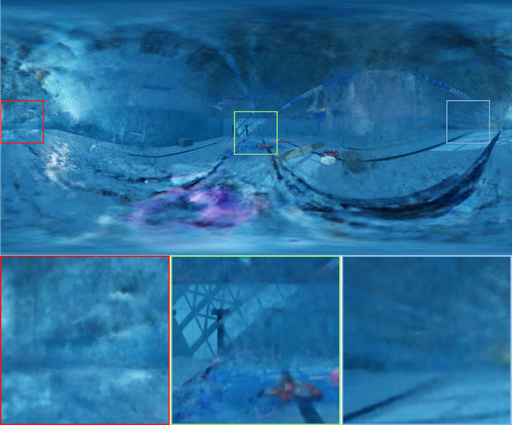} &
\includegraphics[width=.16\textwidth,valign=c]{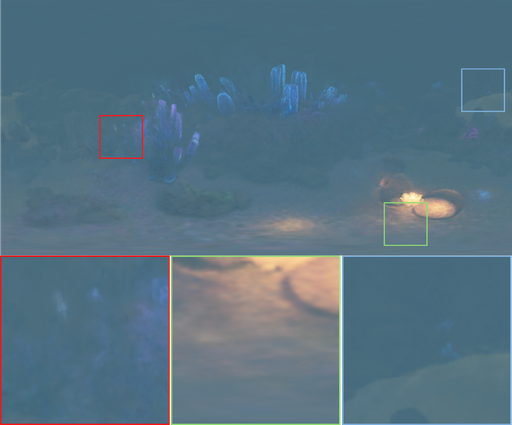} &
\includegraphics[width=.16\textwidth,valign=c]{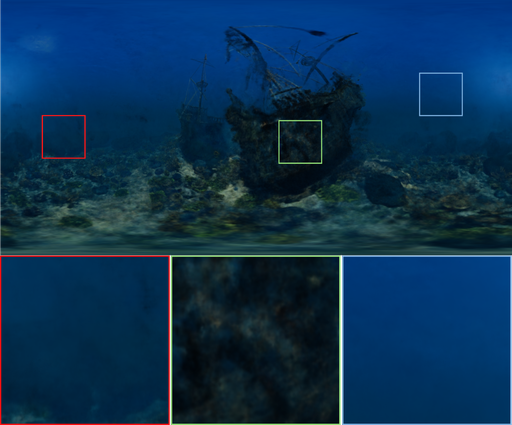} &
\includegraphics[width=.16\textwidth,valign=c]{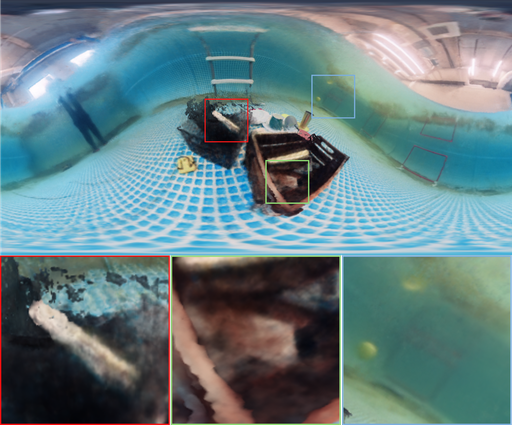} \\
\vspace{1.5pt}
\adjustbox{valign=c}{\rotatebox{90}{\scriptsize\text{2DGS}}} &
\includegraphics[width=.16\textwidth,valign=c]{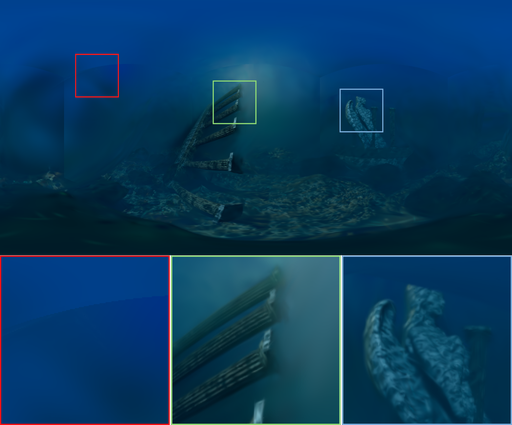} &
\includegraphics[width=.16\textwidth,valign=c]{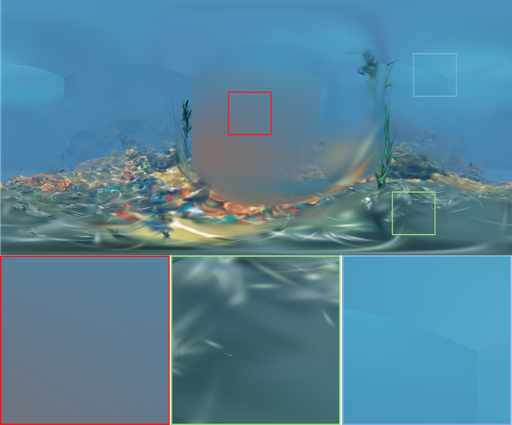} &
\includegraphics[width=.16\textwidth,valign=c]{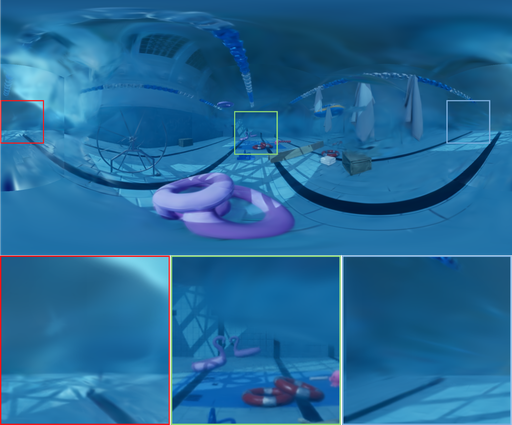} &
\includegraphics[width=.16\textwidth,valign=c]{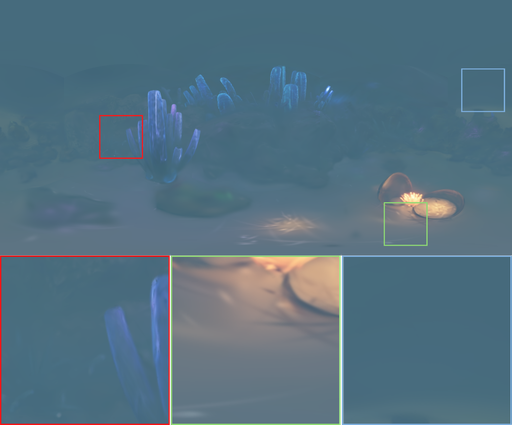} &
\includegraphics[width=.16\textwidth,valign=c]{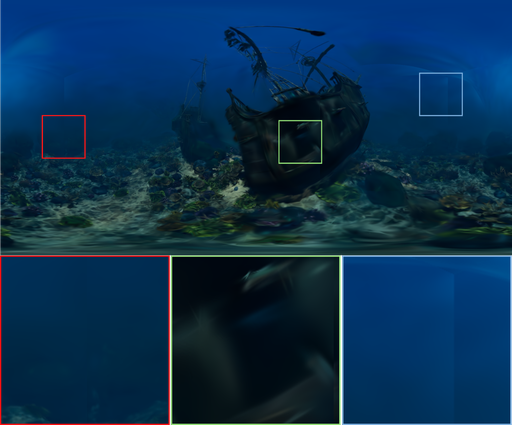} &
\includegraphics[width=.16\textwidth,valign=c]{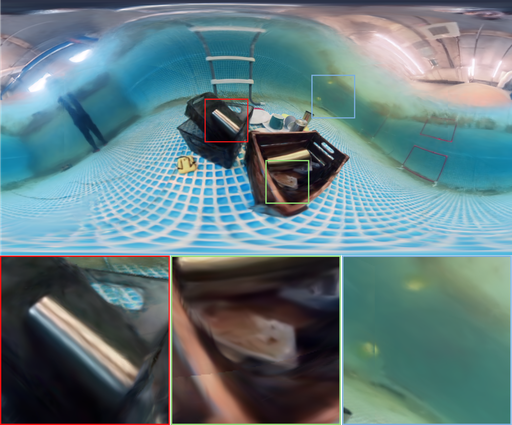} \\
\vspace{1.5pt}
\adjustbox{valign=c}{\rotatebox{90}{\scriptsize\text{Water-Splatting}}} &
\includegraphics[width=.16\textwidth,valign=c]{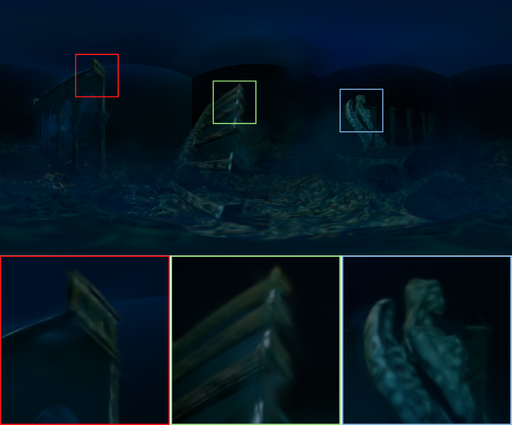} &
\includegraphics[width=.16\textwidth,valign=c]{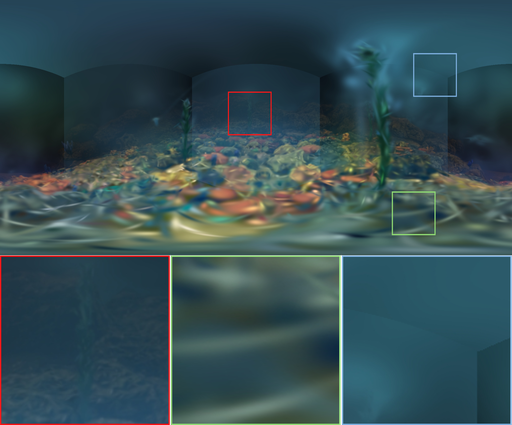} &
\fbox{\parbox[c][.118\textwidth]{.145\textwidth}{\centering N/A}} &  %
\includegraphics[width=.16\textwidth,valign=c]{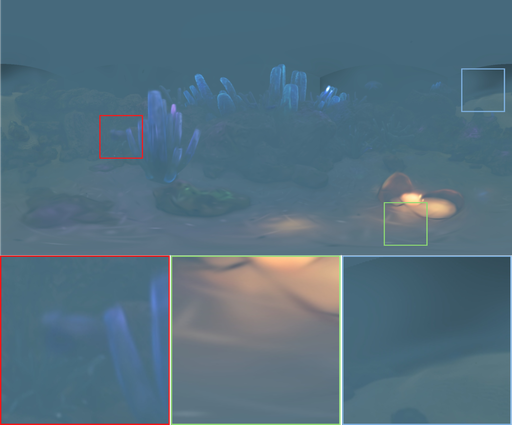} &
\fbox{\parbox[c][.118\textwidth]{.145\textwidth}{\centering N/A}} &  %
\includegraphics[width=.16\textwidth,valign=c]{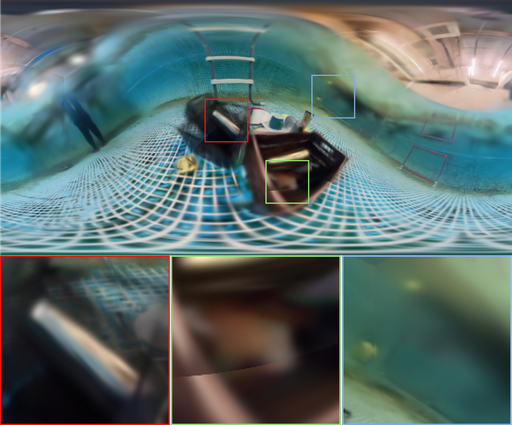} \\
\vspace{1.5pt}
\adjustbox{valign=c}{\rotatebox{90}{\scriptsize\text{UW-GS}}} &
\includegraphics[width=.16\textwidth,valign=c]{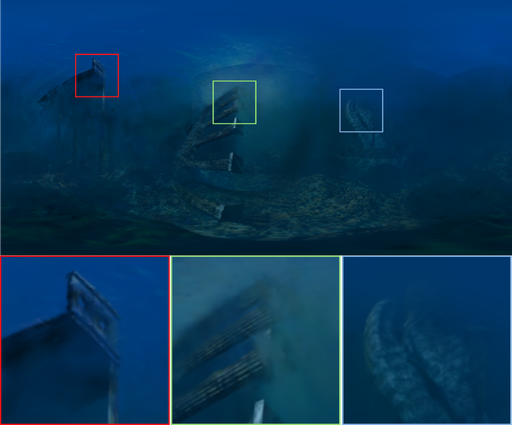} &
\includegraphics[width=.16\textwidth,valign=c]{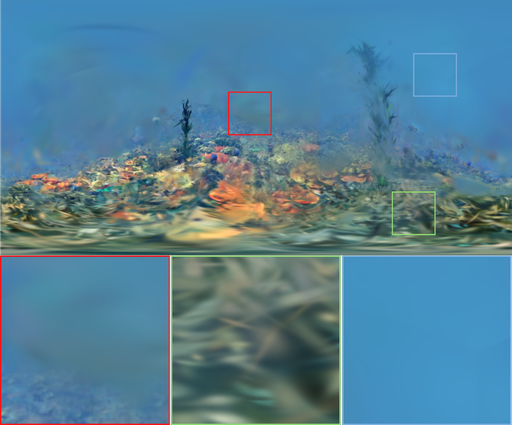} &
\includegraphics[width=.16\textwidth,valign=c]{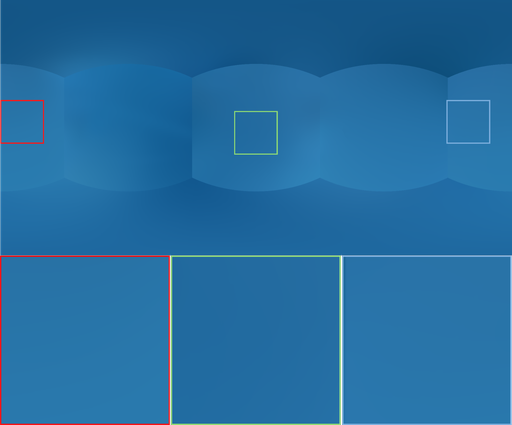} &
\includegraphics[width=.16\textwidth,valign=c]{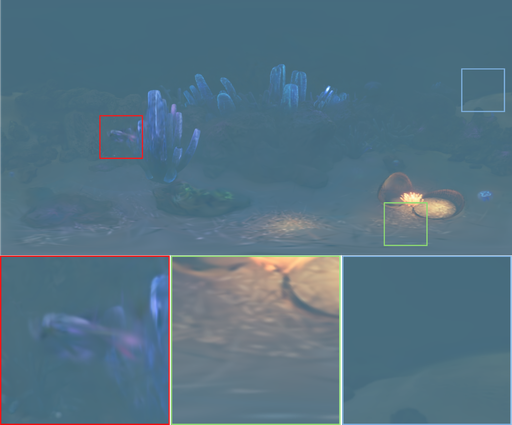} &
\includegraphics[width=.16\textwidth,valign=c]{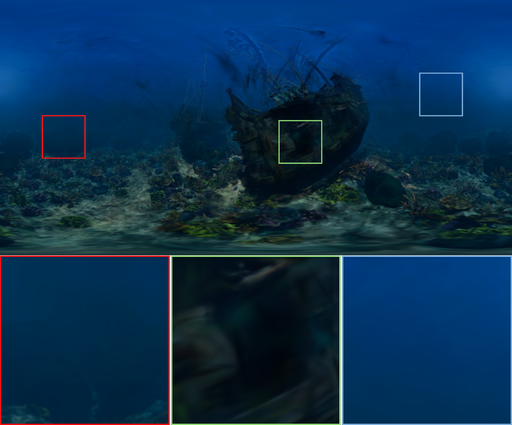} &
\includegraphics[width=.16\textwidth,valign=c]{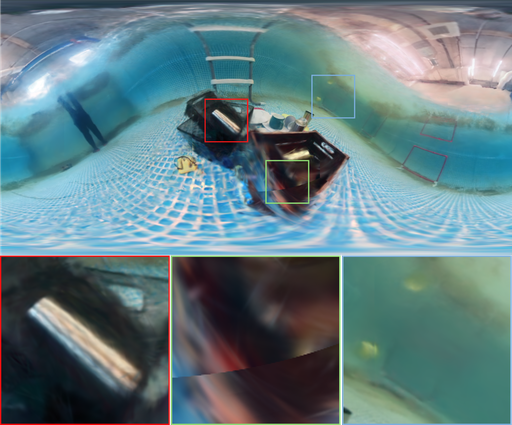} \\
\vspace{1.5pt}
\adjustbox{valign=c}{\rotatebox{90}{\scriptsize\text{ODGS}}} &
\includegraphics[width=.16\textwidth,valign=c]{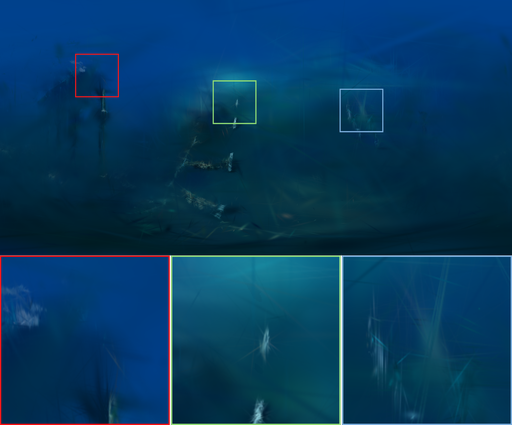} &
\includegraphics[width=.16\textwidth,valign=c]{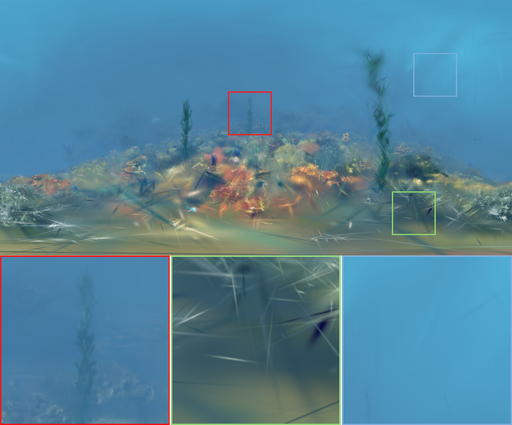} &
\includegraphics[width=.16\textwidth,valign=c]{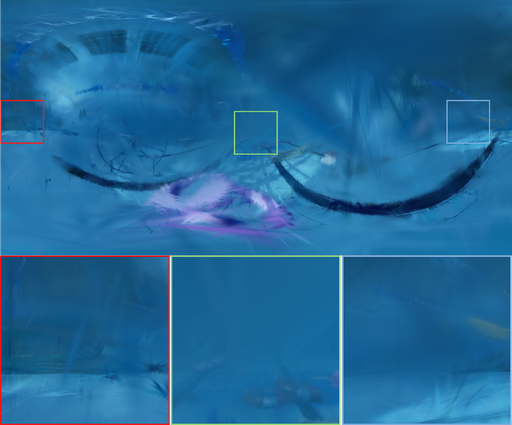} &
\includegraphics[width=.16\textwidth,valign=c]{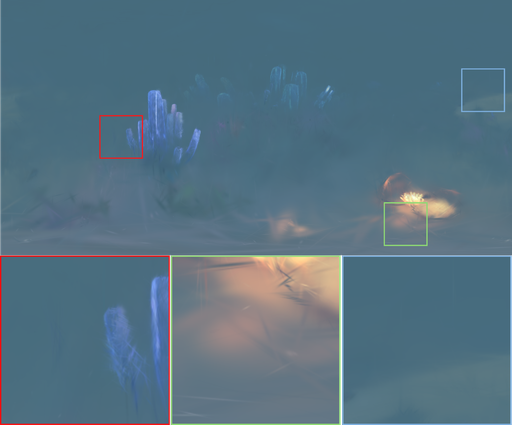} &
\includegraphics[width=.16\textwidth,valign=c]{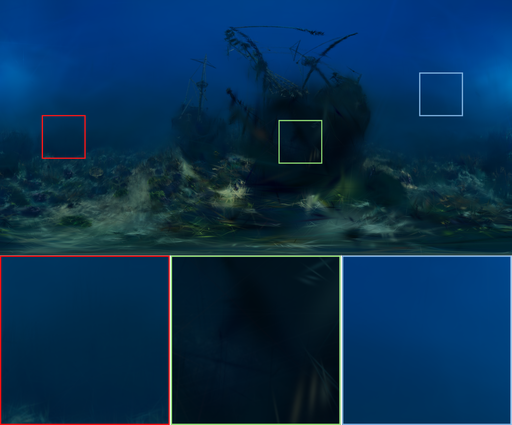} &
\includegraphics[width=.16\textwidth,valign=c]{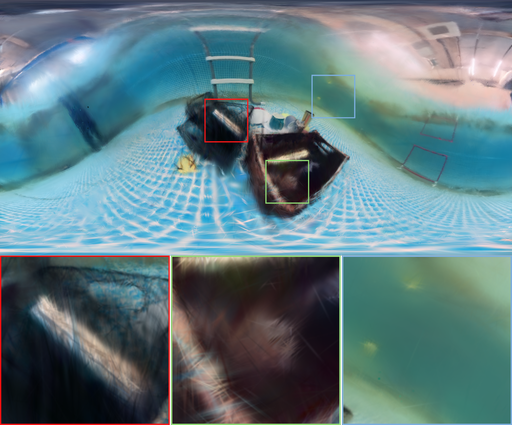} \\
\vspace{1.5pt}
\adjustbox{valign=c}{\rotatebox{90}{\scriptsize\text{OmniGS}}} &
\includegraphics[width=.16\textwidth,valign=c]{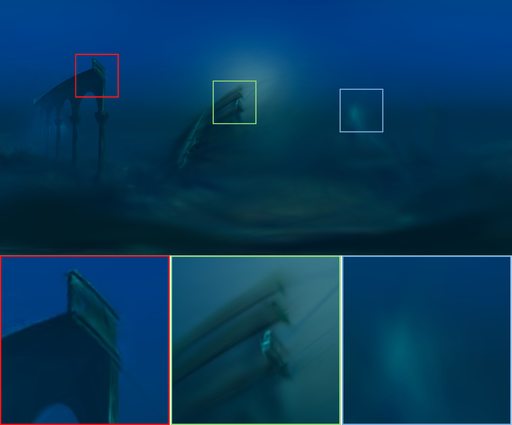} &
\includegraphics[width=.16\textwidth,valign=c]{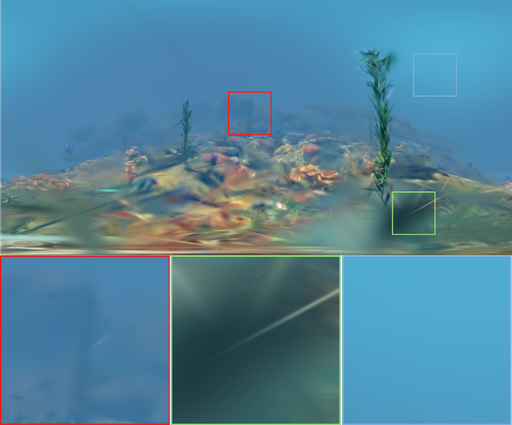} &
\includegraphics[width=.16\textwidth,valign=c]{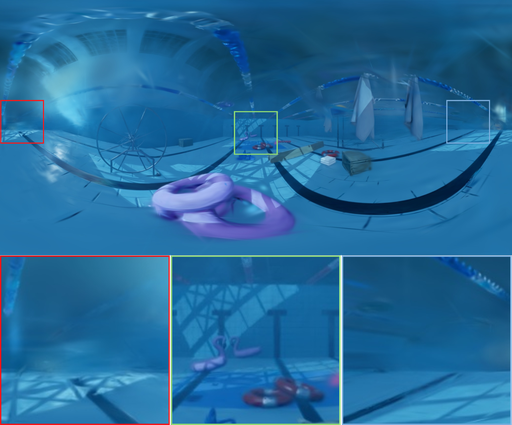} &
\includegraphics[width=.16\textwidth,valign=c]{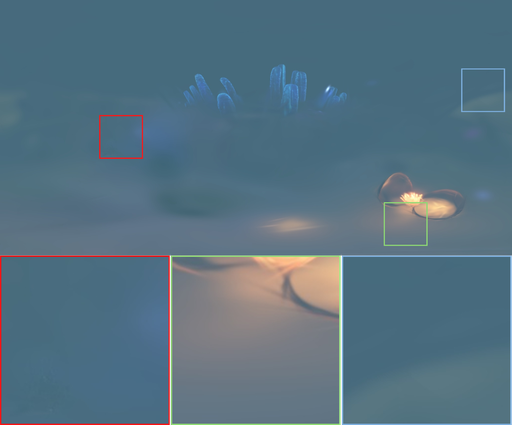} &
\includegraphics[width=.16\textwidth,valign=c]{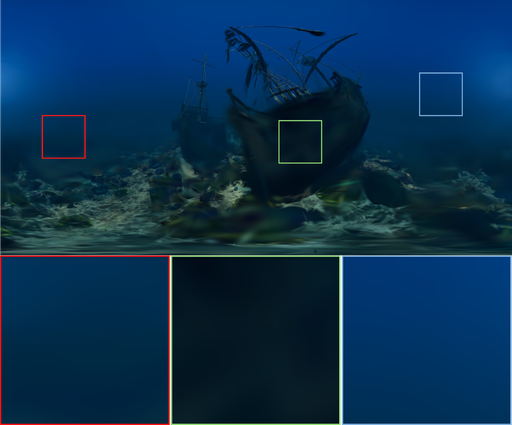} &
\includegraphics[width=.16\textwidth,valign=c]{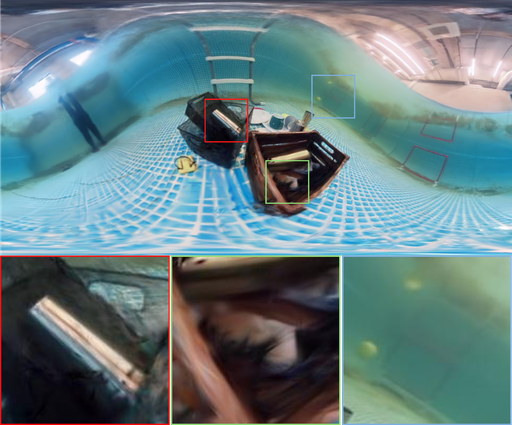} \\
\vspace{1.5pt}
\adjustbox{valign=c}{\rotatebox{90}{\scriptsize\text{Ours}}} &
\includegraphics[width=.16\textwidth,valign=c]{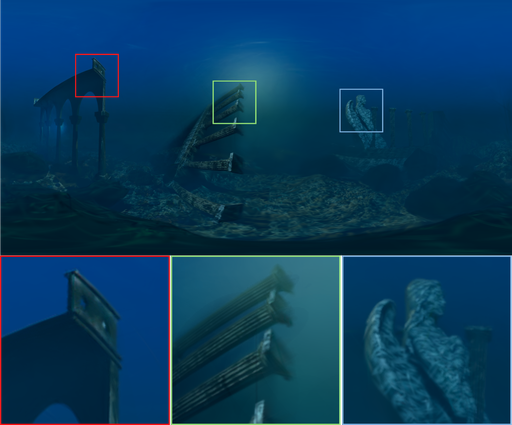} &
\includegraphics[width=.16\textwidth,valign=c]{figs/compare/Ours_deepsea_rgb.png} &
\includegraphics[width=.16\textwidth,valign=c]{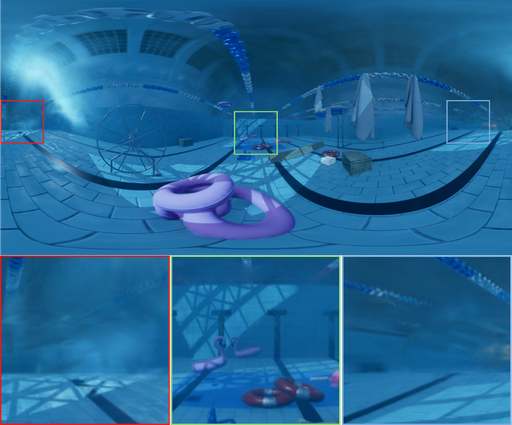} &
\includegraphics[width=.16\textwidth,valign=c]{figs/compare/Ours_rocks_rgb.png} &
\includegraphics[width=.16\textwidth,valign=c]{figs/compare/Ours_ship_rgb.png} &
\includegraphics[width=.16\textwidth,valign=c]{figs/compare/Ours_scene4_rgb.png} \\

\end{tabular}
\vskip-1ex
\caption{More qualitative comparisons on both synthetic and real underwater scenes.}
\label{fig:compare_supple}
\end{figure*}

\begin{figure*}[!htbp]
\centering

\begin{tabular}{@{}c@{\,}c@{\,}c@{\,}c@{\,}c@{\,}}
\vspace{1.5pt}
 & 
\small Attenuation & 
\small Backscatter &
\small Restore &
\small Render \\
\vspace{1.5pt}
\adjustbox{valign=c}{\rotatebox{90}{\scriptsize\text{SeaThru-NeRF}}} &
\includegraphics[width=.24\textwidth,valign=c]{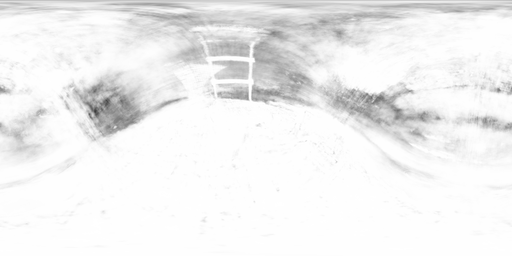} &
\includegraphics[width=.24\textwidth,valign=c]{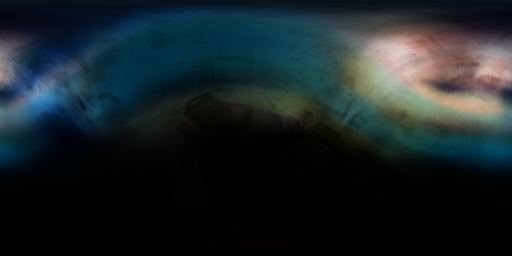} &
\includegraphics[width=.24\textwidth,valign=c]{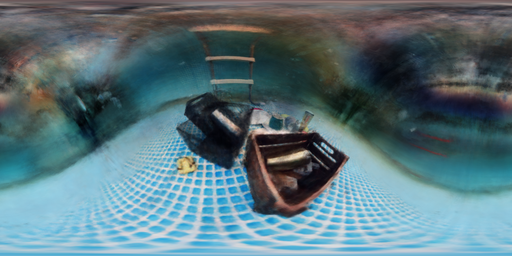} &
\includegraphics[width=.24\textwidth,valign=c]{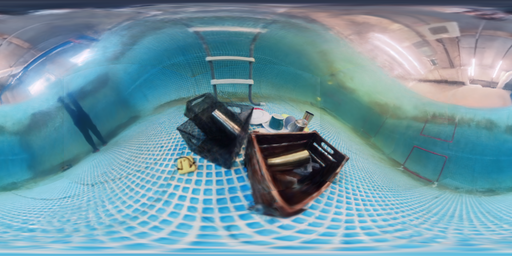} \\
\vspace{1.5pt}
\adjustbox{valign=c}{\rotatebox{90}{\scriptsize\text{SeaSplat}}} &
\includegraphics[width=.24\textwidth,valign=c]{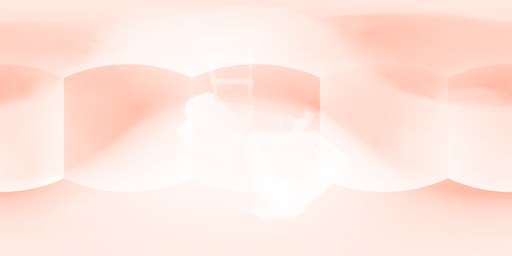} &
\includegraphics[width=.24\textwidth,valign=c]{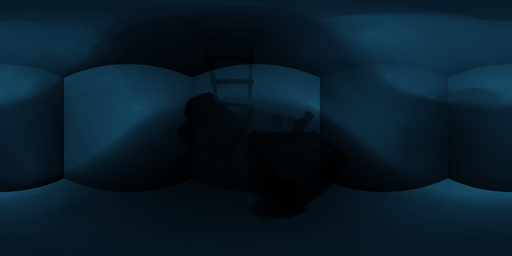} &
\includegraphics[width=.24\textwidth,valign=c]{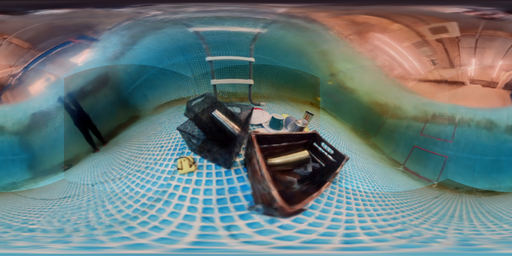} &
\includegraphics[width=.24\textwidth,valign=c]{figs/fig8/SeaSplat_scene4_rgb.png} \\
\vspace{1.5pt}
\adjustbox{valign=c}{\rotatebox{90}{\scriptsize\text{3D-UIR}}} &
\includegraphics[width=.24\textwidth,valign=c]{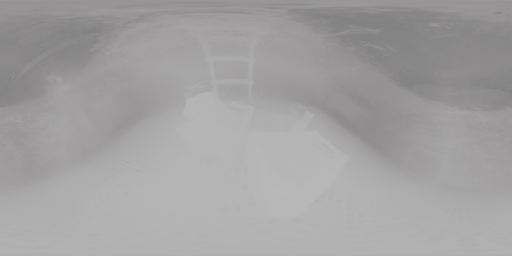} &
\includegraphics[width=.24\textwidth,valign=c]{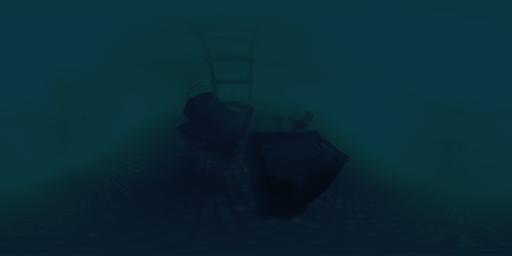} &
\includegraphics[width=.24\textwidth,valign=c]{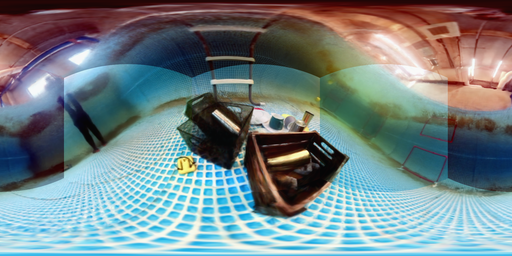} &
\includegraphics[width=.24\textwidth,valign=c]{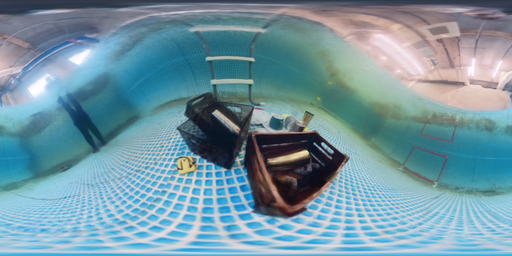} \\
\vspace{1.5pt}
\adjustbox{valign=c}{\rotatebox{90}{\scriptsize\text{Ours}}} &
\includegraphics[width=.24\textwidth,valign=c]{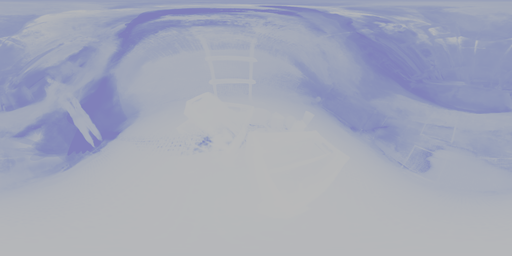} &
\includegraphics[width=.24\textwidth,valign=c]{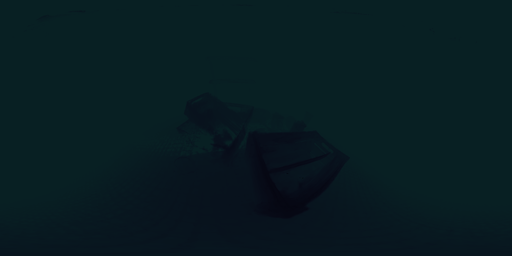} &
\includegraphics[width=.24\textwidth,valign=c]{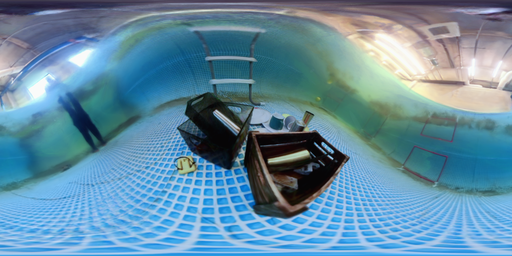} &
\includegraphics[width=.24\textwidth,valign=c]{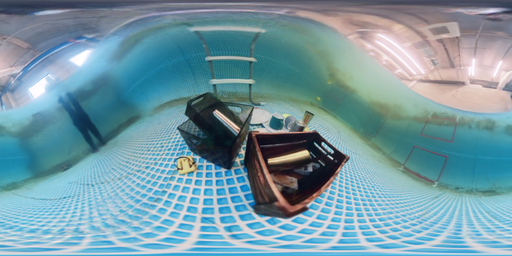} \\

\end{tabular}
\vskip-1ex
\caption{More qualitative comparison of UIFM decomposition results.}
\label{fig:compare_supple_uifm}
\end{figure*}

\end{document}